# Adaptive Compressive Tactile Subsampling: Enabling High Spatiotemporal Resolution in Scalable Robotic Skin

**Authors:** Ariel Slepyan[1]*†, Dian Li (李典)[2]†, Hongjun Cai[1], Ryan McGovern[2], Aidan Aug[2], Sriramana Sankar[2], Trac Tran[1], Nitish Thakor[1,2,3]*

**Affiliations:**

[1]Department of Electrical and Computer Engineering, Johns Hopkins University, 3400 North Charles Street, Baltimore, MD 21218, USA.

[2]Department of Biomedical Engineering, Johns Hopkins School of Medicine, 720 Rutland Avenue, Baltimore, MD 21205, USA.

[3]Department of Neurology, Johns Hopkins University, 600 North Wolfe, Baltimore, MD 21205, USA.

*Corresponding author. Ariel Slepyan: aslepya1@jhu.edu. Nitish Thakor: nitish@jhu.edu

†These authors contributed equally to this work

**Abstract:** Robots, like humans, require full-body, high-resolution tactile sensing to operate safely and effectively in unstructured environments, enabling reflexive responses and closed-loop control. However, the high pixel counts necessary for dense, large-area coverage limit readout rates of most tactile arrays to below 100 Hz, hindering their use in high-speed tasks. We introduce Adaptive Compressive Tactile Subsampling (ACTS), a scalable and data-driven method that dramatically enhances the performance of traditional tactile matrices by leveraging sparse recovery and a learned tactile dictionary. By adaptively allocating measurements to informative regions, ACTS is especially effective for spatially sparse tactile signals common in real-world interactions. Tested on a 1024-pixel tactile sensor array (32×32), ACTS achieved frame rates up to 1,000 Hz, an 18X improvement over conventional raster scanning, with minimal reconstruction error. For the first time, ACTS enables wearable, large-area, high-density tactile sensing systems that can deliver high-speed results. We demonstrate rapid object classification within 20 ms of contact, high-speed projectile detection, ricochet angle estimation, and soft deformation tracking, in tactile and robotics applications, all using flexible, high-density tactile arrays. These include high-resolution tactile gloves, pressure insoles, and full-body configurations covering robotic arms and human-sized mannequins. We further showcase tactile-based closed-loop control by guiding a metallic ball to trace letters using tactile feedback and by executing tactile-only whole-hand reflexes on a fully sensorized LEAP hand to stabilize grasps, prevent slip, and avoid sharp objects, validating ACTS for real-time interaction and motion control. ACTS transforms standard, low-cost, and robust tactile sensors into high-speed systems enabling scalable, responsive, and adaptive tactile perception for robots and wearables operating in dynamic environments.

**One-Sentence Summary:** Data-driven sampling enables a high-resolution, large-area robot skin that remains responsive, robust, and manufacturable.

## Main Text:

## Introduction

Robotics require high-speed tactile arrays for enabling reflexive responses (*1–4*), precise manipulation (*5–7*), and real-time control (*8*, *9*). However, as tactile sensor arrays increase in sensor count, the latency from scanning the array grows (*10–14*). This is because tactile arrays typically use raster scanning or time-division multiple-access (TDMA) methods to measure their pixel values (*9*, *14–17*). Consequently, large arrays with numerous elements can take a significant amount of time to scan fully, resulting in substantial latency and frame rates that rarely exceed 100 Hz (*18*) **(see supplemental table S1)**, when the minimum desired sampling rate for robotic control is 1 kHz (*19*).

To mitigate these scanning delays, researchers have typically turned to more powerful processors, including custom field programmable gate array (FPGA) solutions (*18*, *20*, *21*), to accelerate scanning times. Despite these advancements, such current solutions have not scaled well and are constrained by the inherent linear limitations of raster scanning, where improvements in processing speed only offer linear gains. Another class of solutions that researchers have explored is based on embedded computing and delta encoding (*13*, *22–24*). These solutions embed circuits within or near the tactile sensing pixels that continuously monitor the sensor for significant changes in the measurements. When a significant change is detected, only that change is transmitted over a shared communication bus. While this approach enables low latencies and high-speed tactile data transmission, it requires substantial hardware modifications to the robot's skin due to the integration of integrated circuits (ICs) near the sensor. This introduces several drawbacks, including limitations on sensor density, high fabrication costs, reduced sensor flexibility, and compromised durability and robustness. Furthermore, this method restricts the use of exotic sensing materials or structures that are challenging to integrate with ICs.

In other fields, such as optical imaging, ultrasound measurements, and neural recording, similar latency challenges have been addressed using compressed sensing techniques (*25–34*). Compressed sensing (CS) is a signal processing technique that leverages patterns in data to reconstruct a signal with high fidelity from a few measurements of that signal. By taking fewer measurements, CS systems have been used to develop high-speed imagers (*35*) and lower-power and lower-cost sensing devices (*36*).

Despite the potential of CS-based techniques, they have not yet been translated to tactile sensing due to fundamental differences between tactile skin and monolithically integrated devices like cameras. Typical CS systems require pixel-wise control to perform summations of pixels with programmed weights, necessitating substantial electrical traces and circuits. However, tactile skin is inherently a distributed process designed to cover large surfaces using flexible or stretchable materials (*12*). This imposes technical limits on wiring density, requires the use of dedicated multiplexers, and limits data rates by analog-to-digital converters (ADCs) – all severely challenging the requirements of a traditional CS system.

Because of these challenges, limited work has been done in compressed tactile sensing, demonstrated only through simulations (*37–41*). One experimental study used only a few sensors and required individual pixel connections, which is impractical for robot skin at scale (*42*, *43*). Generally, both simulation and experimental works treat the tactile problem like a classic image problem.

However, despite the challenges in implementation, from a data compressibility perspective, tactile data is very well fitted for compressed sensing. This is because tactile data has significant repetitive data patterns and high spatiotemporal sparsity (*44*, *45*).

With all this in mind, we developed a compressive tactile sensing system that does not attempt to change the hardware of a typical tactile sensing system through embedded computing, but rather changes the sampling pattern and method of reconstructing tactile signals. We present ACTS, adaptive compressive tactile subsampling: a subsampling-based method designed specifically for tactile sensing systems that leverages the spatial sparsity of tactile signals to quickly and efficiently 'search' for tactile activity while taking very few measurements and reconstructing full tactile frames using a learned tactile dictionary.

To the best of our knowledge, the most closely related work is a subsampling paper (*46*) that explored the use of 'smart sampling' to improve the temporal resolution of tactile sensors. However, their strategy is inefficient for tactile data as it relies on random sampling and probing measurements that travel towards the direction of pressure; with poor adaptation to multi-touch or complex shapes and contacts.

This work is transformative because it bridges the gap between traditional tactile sensor array designs – robust, thin, flexible, high-density, and scalable to large areas – and the high-speed performance required for next-generation robotics. Historically limited by slow readout rates, these arrays can now achieve the spatiotemporal resolution necessary for robotic reflexes and closed-loop control in dynamic, unstructured environments. This advancement equips robots with the ability to rapidly classify and react to contacts across their entire surface, detect and localize object shapes for precise manipulation, and track high-speed moving objects. By enabling comprehensive tactile coverage, ACTS paves the way for robots to perform more lifelike interactions. The firmware-based nature of ACTS means existing robotic systems can be upgraded without hardware modifications, accelerating the adoption of high-performance tactile sensing across the field. Moreover, by reducing the computational load through efficient subsampling, ACTS supports energy-constrained platforms, such as wearable robotics or mobile systems. Its Arduino-compatible, open-source implementation ensures accessibility for researchers, developers, and industries alike.

In this article, we explore the applications, benefits, and trade-offs of ACTS in tactile sensor arrays, showcasing its potential to revolutionize robotic perception and interaction. Our primary "adaptive" method, inspired by binary search, is compared with alternative approaches like uniform and random subsampling, demonstrating robust performance in diverse settings.

Our adaptive approach rapidly hones in on areas of contact, often providing sufficient information for many tasks directly from its search pattern, with dictionary-based sparse recovery being optional unless measurement budgets are constrained.

Our testing is conducted with a 1024-pixel tactile array (32x32) and reconstruction error and classification accuracy are assessed during interactions with a library of 30 everyday objects. We demonstrate that our implementation can generalize to new and untrained tactile stimuli, enhance rapid contact detection in high-speed interactions, improve contact estimation with deformable materials, estimate ricochet angles, and accurately sense impacts of high-speed projectiles and dynamic collisions. Additionally, ACTS seamlessly adapts to various sensor shapes and configurations, from high-density gloves and insoles to full-body coverage for human-sized mannequins or robotic platforms. By advancing tactile sensing, ACTS not only addresses critical needs in robotics but also opens doors to a wide array of applications in wearable systems, intelligent interfaces, and any technology requiring large-scale, high-speed sensing.

## Results

### *Compressive Tactile Subsampling*

Compressive Tactile Subsampling reconstructs tactile data with simultaneously high temporal and high spatial resolution for dynamic interactions (**Fig. 1**). To sample the dense tactile array containing many sensors quickly, only several sensors are measured in a given tactile frame, dramatically improving tactile frame rates. These subsampled measurements are then used to reconstruct the full tactile signal using knowledge of the spatial patterns in tactile data, through a learned tactile dictionary. Combined, compressive tactile subsampling allows a robotic system to obtain high spatiotemporal tactile sensing at low data throughputs without demanding very fast or powerful hardware (**Supplemental S2**). Furthermore, any conventional resistive sensor array with a conventional readout circuit can be used, and current tactile systems can be upgraded to obtain the new compressive tactile subsampling features.

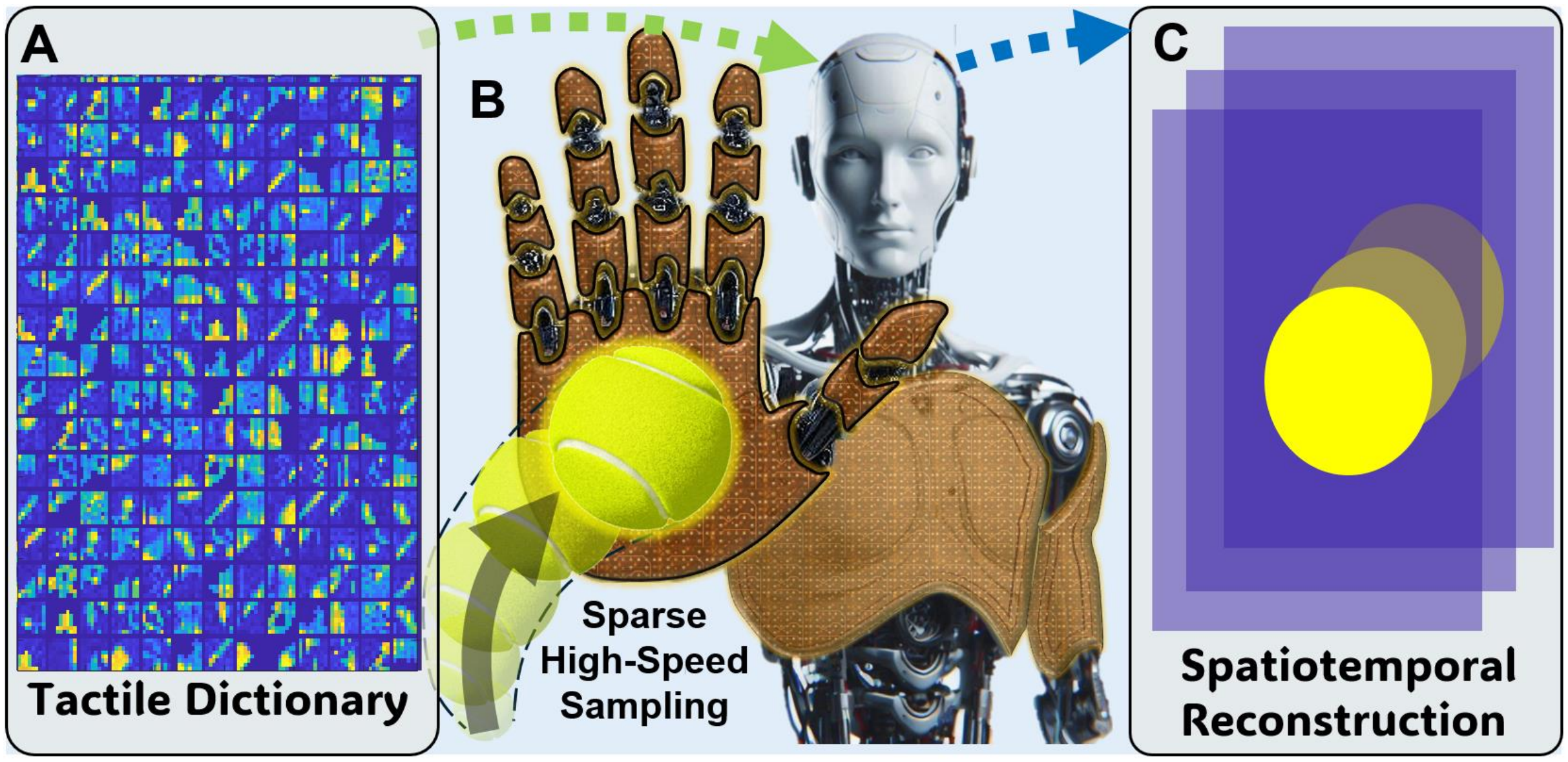

**Fig. 1. High-Speed Compressive Tactile Subsampling. (A)** Robotic sensing system has prior knowledge of patterns in tactile data through a learned tactile dictionary. **(B)** Sparse compressive subsampling is used to measure tactile data at high frame rates. **(C)** High spatiotemporal resolution tactile data is reconstructed based on the compressed measurements using the known tactile dictionary and sparse recovery.

Compressive tactile subsampling relies on tactile data containing reliable spatial patterns such that a tactile frame with $N$ pixels can be expressed as a linear combination of $S$ patterns, where $S \ll N$ $S \ll N$. Equivalently, if the tactile patterns form a dictionary $\psi$, tactile signal $x$ can be approximately represented as:

$$x \approx \sum_{i=1}^{S} \alpha_i \psi_i \tag{1.1}$$

where $\alpha$ represents the contribution of each dictionary element in the signal.

When $\psi$ is an overcomplete dictionary containing $K$ many tactile spatial patterns, the dictionary allows for a sparse representation $x_s$ of the tactile signal

$$x \approx \psi x_s \tag{1.2}$$

Representation of tactile data as a sparse vector allows for the use of sparse recovery algorithms to estimate $x_s$ from relatively few measurements, as a classical compressed sensing problem (**Fig. 2**).

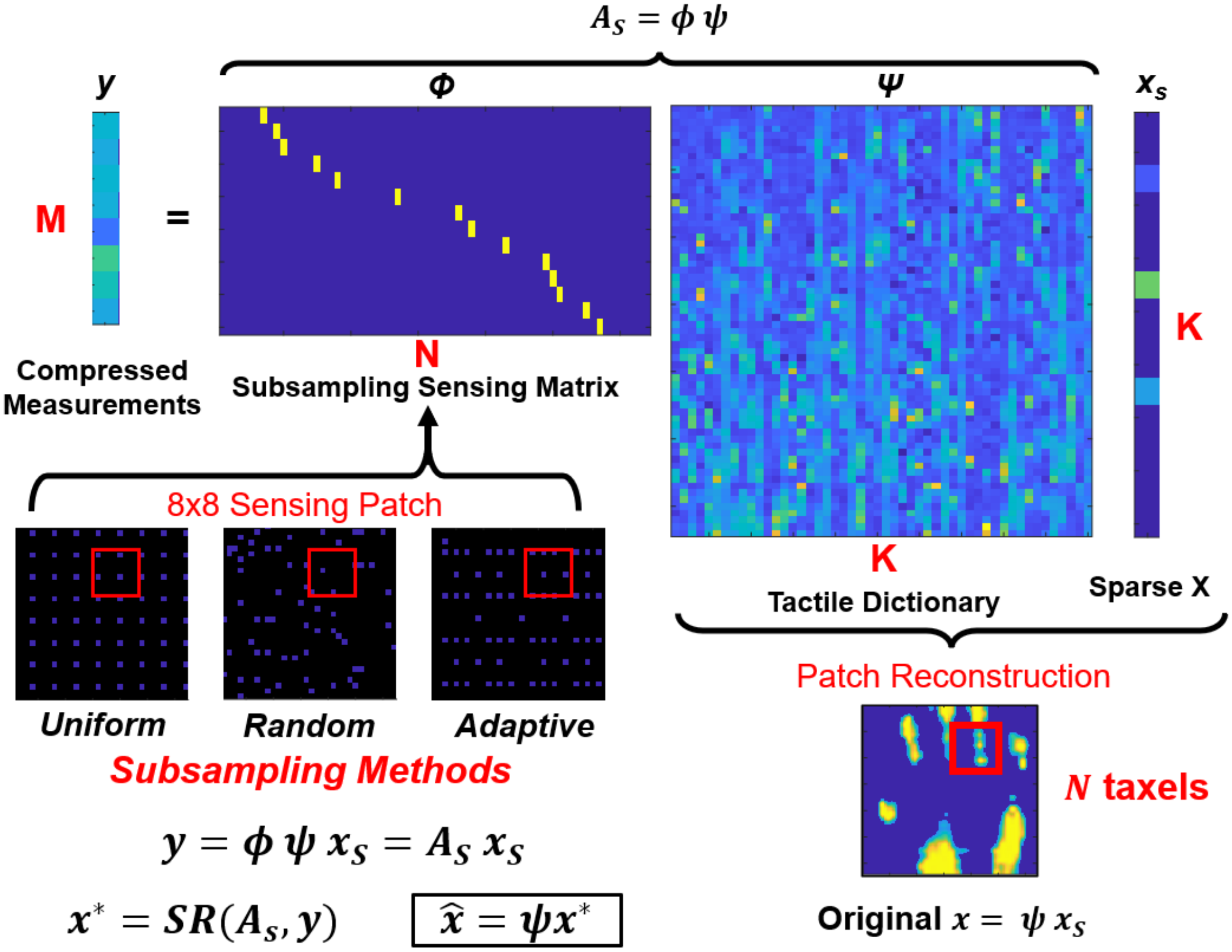

**Fig. 2. Compressive Tactile Subsampling Formulation.** Tactile data $\boldsymbol{x}$ with N taxels is represented as a sparse combination $\boldsymbol{x_S}$ of K dictionary atoms $\boldsymbol{\psi}$. Subsampling matrix – either uniform, random, or adaptive – takes measurements of $\boldsymbol{x}$ using the measurement scheme in $\boldsymbol{\phi}$, to obtain $\boldsymbol{y}$, which has M << N number of measurements. Sparse recovery (SR) is used to calculate a reconstruction of the original tactile data $\hat{\boldsymbol{x}}$ using measurements $\boldsymbol{y}$ and knowledge of the sensing matrix $\boldsymbol{\phi}$ and dictionary matrix $\boldsymbol{\psi}$.

To obtain these few measurements, a sampling strategy $\phi$ is applied. For compressive tactile subsampling, each measurement contains the data from a single pixel, hence the rows of sensing matrix $\phi$ are one-hot and binary, with only one coefficient being non-zero with a value of 1. To achieve benefits on framerate, the number of measurements $M$ should be much less than the number of total sensing pixels $N$.

In this work we explore three different sensing scheme styles: uniform subsampling, random subsampling, and adaptive subsampling. Uniform subsampling chooses pixels to sample following a uniform pattern. Random subsampling chooses pixels at random to sample. Adaptive sampling recursively samples sections of the sensor array until notable pressures are encountered. The number of measurements $M$ is experimentally varied. Overall, the sensing process can be described as:

$$y = \phi x = \phi \psi x_s \tag{1.3}$$

where $y \in \mathbb{R}^M$ $y \in \mathbb{R}^M$, $\phi \in \mathbb{R}^{MxN}$ $\phi \in \mathbb{R}^{M \times N}$, $x \in \mathbb{R}^N$ $x \in \mathbb{R}^N$, $\psi \in \mathbb{R}^{NxK}$ $\psi \in \mathbb{R}^{N \times K}$, $x_s \in \mathbb{R}^K$ $x_s \in \mathbb{R}^K$. Although $M < N$ causes the problem to be ill-posed, it has been shown that the problem can be uniquely solved with high probability if $x_s$ is a S-sparse vector and $M \geq 2S$ (*47*). Under these conditions, and if the sensing matrix is sufficiently incoherent with the dictionary matrix (*48*), sparse recovery algorithms can be used to solve for $x_s$. Then the original signal can be reconstructed using the tactile dictionary: $x = \psi x_s$. In this work a hardware-friendly greedy pursuit algorithm, which we call FastOMP, is used for real-time sparse recovery (*49*).

Compressive tactile subsampling was implemented using a conventional zero-potential resistive sensor array readout circuit (*50*), and a commodity microcontroller (MCU, Teensy 4.1, *PJRC*). The MCU is programmed to execute the subsampling scheme by controlling the digital input rows to the sensor array and multiplexing the readout columns according to the current sampling scheme (**Fig. 3**). The process is repeated until $M$ measurements are obtained.

The MCU then reconstructs the full tactile signal $x$ using FastOMP from the measurements $y$ and the known $\phi$ and $\psi$. The tactile sensor array is arranged with 32 rows and 32 columns forming 1024 tactile sensing pixels (taxels). An inverting amplifier is added after the summing amplifier to bring the output voltage into the positive readable range of the ADC (**Fig. 3**).

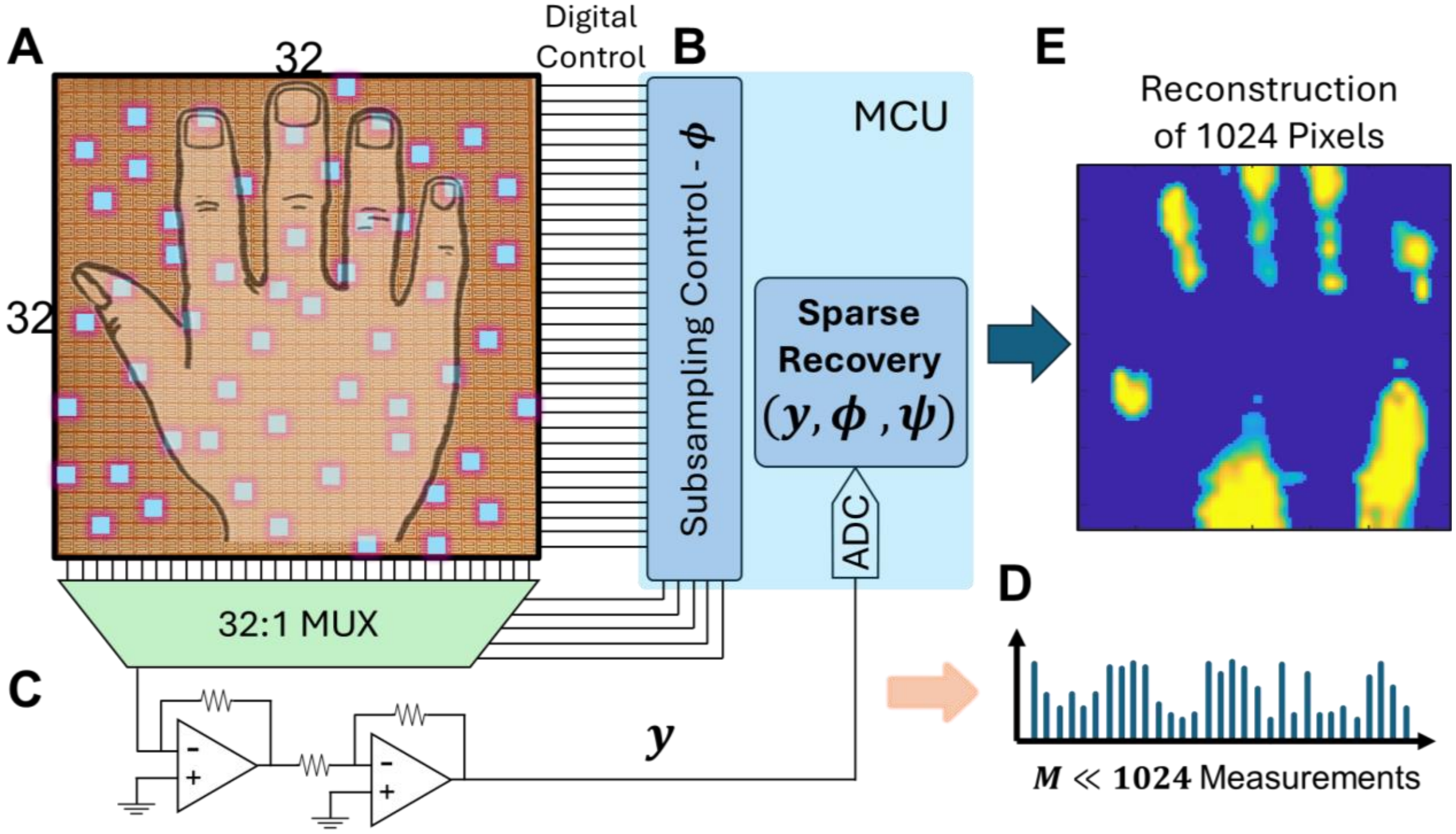

**Fig. 3. Compressive Subsampling Hardware Implementation. (A)** 32x32 tactile sensor array developed on a flex-PCB with a piezoresistive Velostat layer. **(B)** Microcontroller (MCU) performs compressive subsampling of the tactile array by programmatically controlling digital inputs to the rows of the sensor array and multiplexing the readout columns of the sensor array. **(C)** Zero-potential summing circuit is used to accurately readout the resistance of the sensing element and a inverting amplifier circuit is then used to rectify the output voltage into the range of the ADC. **(D)** M number of measurements are taken by the ADC where M is much less than the total number of pixels, 1024. **(E)** A reconstruction of the tactile data is computed using sparse recovery.

### *Rapid Tactile Reconstruction and Classification of Daily Objects*

The different sampling strategies were tested under tactile interactions with a library of 30 daily objects and 3D printed shapes (**Supplement S13).** The objects were indented by a robotic arm (UR5e from Universal Robots) into the compressive tactile sensor array while the MCU varied the measurement level and sampling scheme, collecting 10 indentation trials for each test case (**Fig. 4A**).

The frame rate (FPS) of the compressive tactile sensor is inversely proportional to the measurement level with the relationship $FPS = 55936 \frac{samples}{\text{sec}} \Big/ M$ for all three sampling schemes, with $M \approx 55$ yielding approximately 1000 FPS (**Fig. 4D**).

Reconstruction of the full tactile frame is computed using the tactile dictionary (**Supplement S5**). Comparison to other dictionaries is shown in **Supplement S8**. Examples of the corresponding measurements and reconstructions are shown in **Fig. 4B** for the adaptive sampling method and **Fig. 4C** for the random sampling method of the "X" shape. The support accuracy of the tactile reconstruction is calculated to evaluate the quality of each reconstruction at different measurement levels and is compared to a naïve reconstruction using linear interpolation. The support accuracy captures the accuracy of selected pixels and generally reflects the shape of the indented objects. The average support accuracy across all objects is shown in **Fig. 4E** for each method. The objects in the dataset generally occupied less than 50% of the sampled pixels on average and hence overall the adaptive sampling method yielded the highest support accuracy at all levels of measurement. However larger objects were better reconstructed using random and uniform sampling than small objects at low measurement levels (**Supplement S6**). Full confusion matrices are shown in **Supplement S7**.

A classification test using Sparse Representation-based Classification (SRC) was conducted to evaluate the different subsampling methods in object detection at different measurement levels. Adaptive sampling is the most efficient for classification and can classify the objects in the library with 99% accuracy using M=88 measurements (636 FPS), and 90% accuracy at M=56 (999 FPS) (**Fig. 4F**).

To showcase one of the utilities of the higher frame rate sampling methods, a rapid classification test was conducted to classify the indented objects immediately after first contact (**Fig. 4G**). At only 20 ms after first contact, the adaptive method (M=64) can classify with 88% accuracy, while uniform (M=256) and random (M=256) sampling achieve 71% and 68% accuracy respectively. Raster scan (M=1024) can only achieve 51% accuracy in this first contact window.

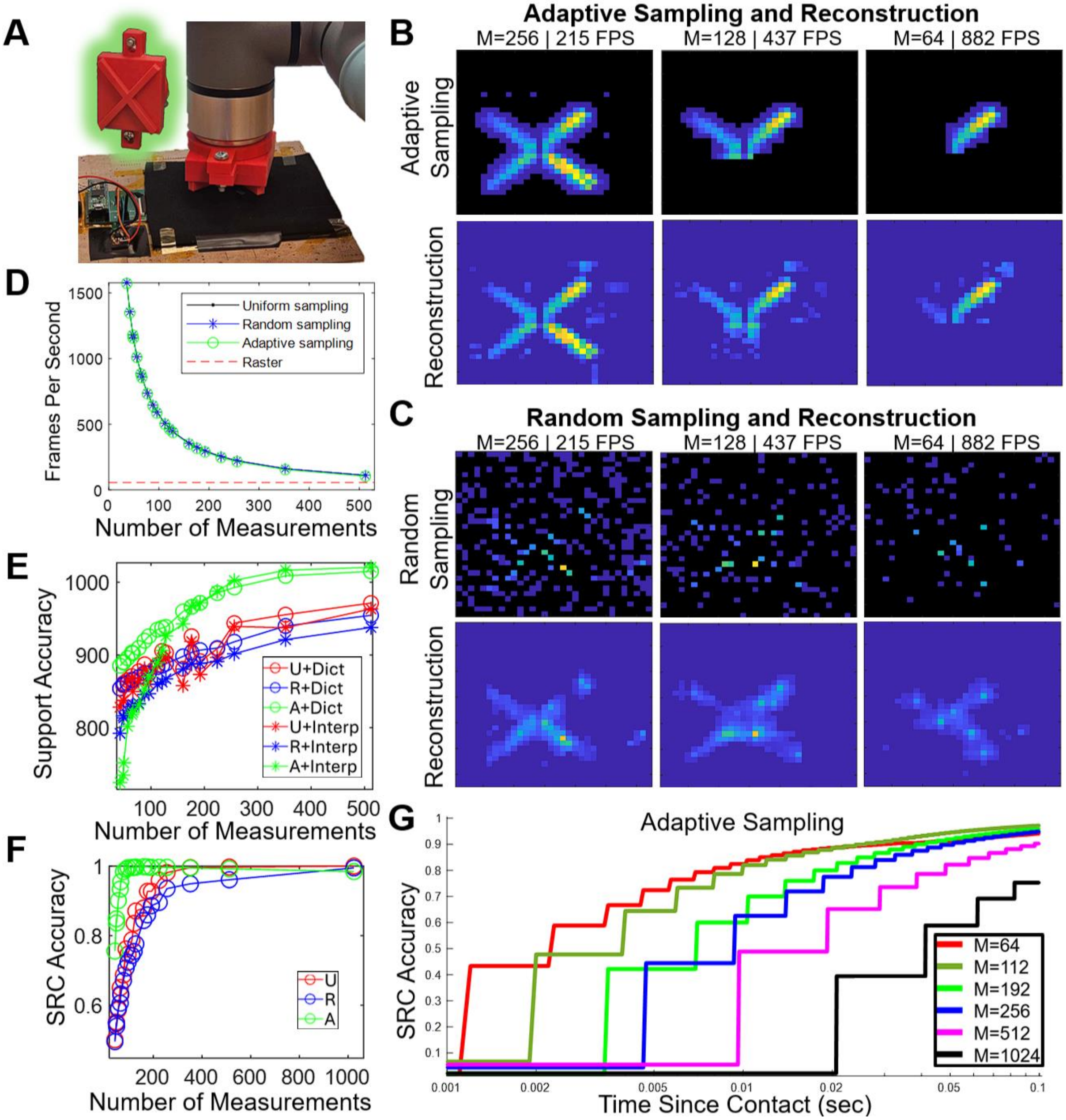

**Fig. 4. Compressive Subsampling Indentation Evaluation. (A)** 30 different objects, such as "X", are attached to the end effector of the UR5e Robotic Arm and are repeatedly indented into the tactile sensor array while the different sampling schemes (Uniform, Random, and Adaptive) are tested at different measurement levels. **(B)** Adaptive sampling and Reconstruction of the "X" at three different measurement levels. **(C)** Random Sampling and Reconstruction of the "X" at three different measurement levels. For **(B)** and **(C)** black pixels are not measured, and lighter colors represent higher pressures. **(D)** Frame rate (FPS) of three sampling methods versus measurement level, red-dotted line shows the faster scan frame rate of 55 FPS. **(E)** Average support accuracy across all objects of different sampling and reconstruction methods versus measurement level. **(F)** Average SRC object classification accuracy across all objects of different sampling methods versus measurement level **(G)** Average classification accuracy after first contact versus measurement level using Adaptive sampling method.

### ***High-Speed Projectile Tracking***

The enhanced frame rate of the compressive tactile sensor enables it to accurately capture the dynamics of high-speed collisions and interactions. To demonstrate this, tennis balls were bounced onto the sensor array from table height, where contact with the sensor lasted approximately only 8.7 ms. Using the adaptive sampling method, the compressive tactile sensor can track the tennis ball for an average of 11.6 frames per bounce at M=42. Larger measurement levels take more time to capture and hence the projectile is seen for fewer frames. For example, raster scan M=1024 usually cannot see the bounce, and can see the projectile for on average only 0.2 frames per bounce (**Fig. 5B**). Furthermore, capturing more frames over the bounce shows more gradual changes in the force of the bounce during the collision (**Fig. 5C**). This captures the interaction more smoothly as shown in examples **Fig. 5D** where M=64 has a typical tactile profile (plateau shape), but M=512 can see the bouncing event during only 1 measurement (**Video V1**). A rubber ball was also bounced off the sensor at a shallow angle to perform a ricochet (**Fig. 5E**). The enhanced spatiotemporal resolution enables tracking the evolution of the center of pressure over time (**Fig. 5F**). The progression of center of pressure can be used to predict the ricochet angle that the ball was originally thrown at (**Fig. 5G**). Other angles are shown in **Video V2**. The enhanced spatiotemporal resolution can also handle the detection of fired foam bullets, but with fewer captured frames (**Video V3**).

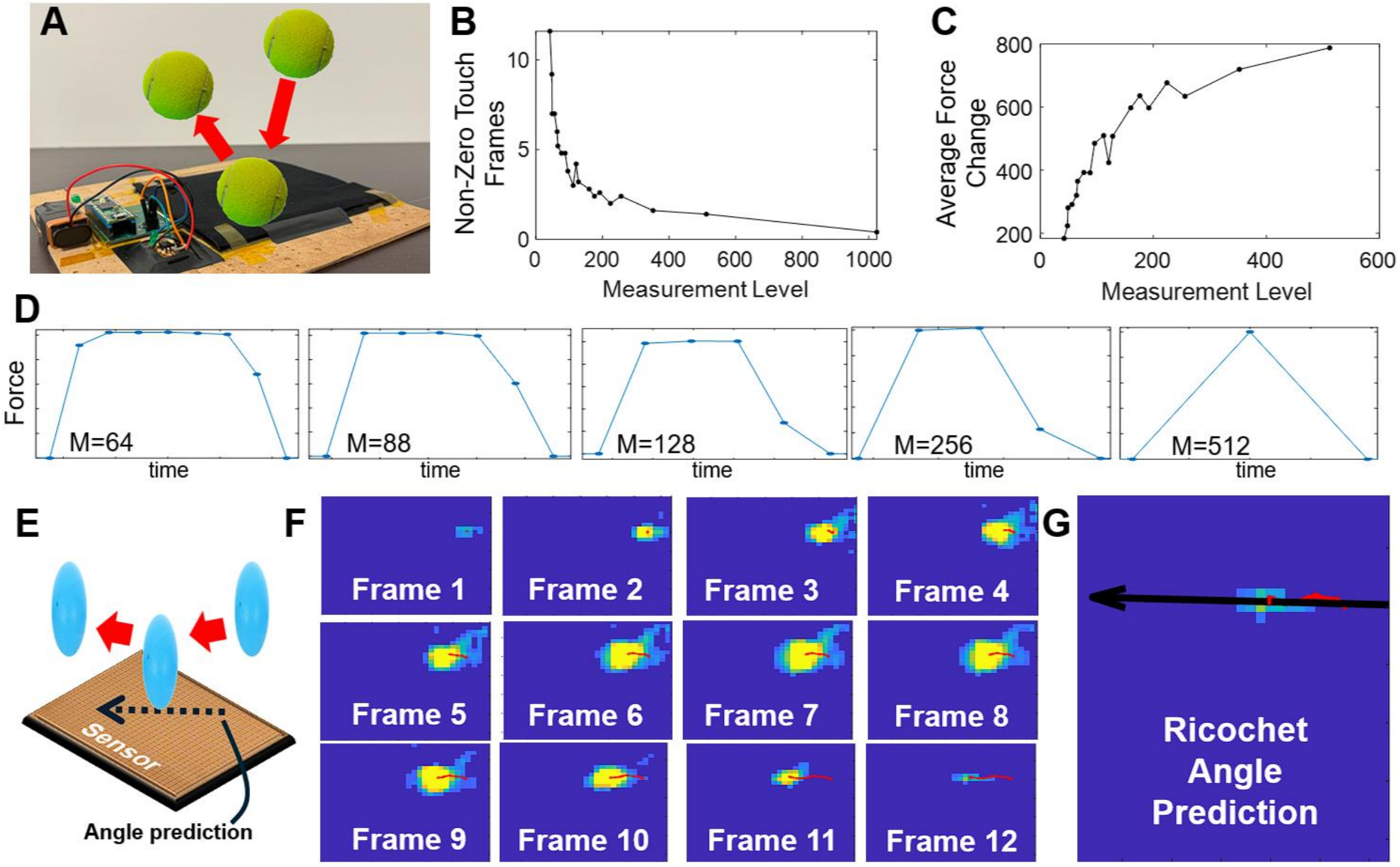

**Fig. 5. High speed projectile tracking. (A)** Tennis ball bouncing off tactile sensor array with ~8.7ms contact time. **(B)** Number of detected tennis ball frames for different measurement levels. **(C)** Average change of force for pressure readings during tennis ball bounce at different measurement levels, smaller values denote smoother pressure tracking. **(D)** Measured force versus time during projectile impact at different measurement levels. **(E)** Rubber ball ricochetting off tactile sensor array at an angle. **(F)** Recorded frames during the ricochet with evolution of center of pressure plotted over time with a red line. **(G)** Estimated ricochet angle by summing the total center of pressure changes plotted as black line.

### *High-Resolution Spatiotemporal Deformation Sensing*

The simultaneous high-speed and high-density of the compressive tactile sensor enables spatiotemporal tracking of deformable objects during indentation. Deformable objects rapidly change their profile upon collision with the sensor which is difficult to measure. Compressive tactile sensing allows for high resolution spatiotemporal tracking of dynamic contacts.

To demonstrate this, we bounced 3 deformable objects into the compressive tactile sensor: “tennis ball”, “silicone block”, “balloon”. The silicone block was bounced on its corner edge. The “tennis ball” and “silicone block” were measured using adaptive sampling, and the “balloon” was measured using random sampling because of its large contact area. The bouncing indentations were repeated for different measurement levels and the collected measurements were reconstructed and the outlines of the shapes in each frame were calculated. The radius of reconstruction for the balloon was calculated to approximate the outline. The outlines of the shapes are plotted over time where lower M values lead to greater resolution in time of progression of the outline during the indentation (**Fig. 6**). Small M values lead to faithful 3D spatiotemporal reconstructions and large M values lack temporal detail and can fail to show the spatiotemporal deformations in the “tennis ball” and “silicone cube” (**Video V6**).

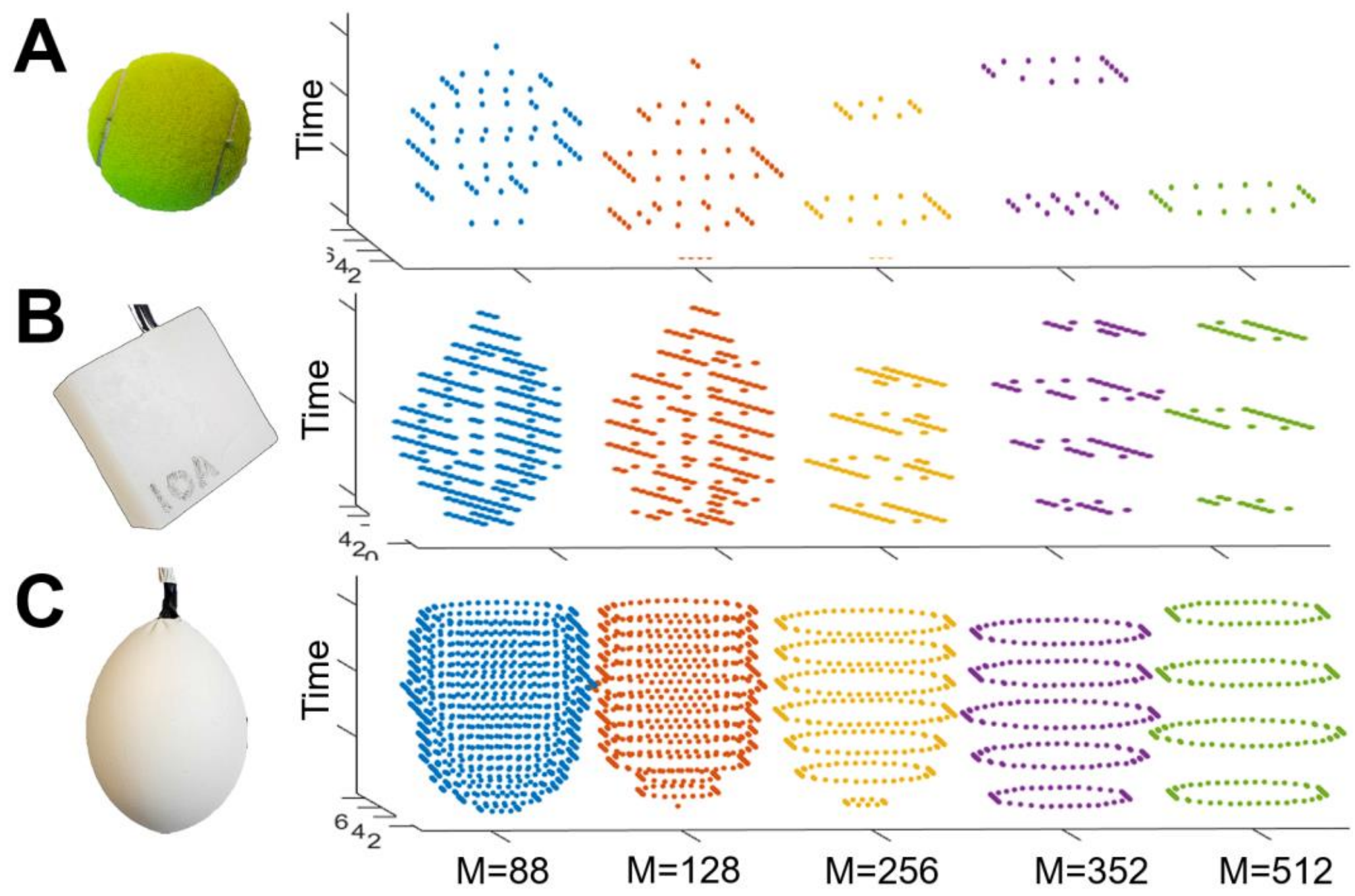

**Fig. 6. High-speed shape estimation of deformable objects impacting sensor**. The object outline during each measured frame is plotted over time (z-axis), to show the temporal progression of the deformable contact captured at different measurement levels. **(A)** Tennis ball impacting sensor for 8 ms. **(B)** Silicone block impacting sensor on its corner edge for 10ms. **(C)** Balloon impacting sensor for 30 ms.

### *Large-Area Spatiotemporal Sensing*

Because our approach leverages conventional resistive sensors, we can flexibly and scalably cover extensive surfaces while maintaining durability and enhancing temporal resolution. Unlike sensors with embedded electronics, which may be prone to damage during intense interactions, our sensor array can endure rough handling and is well-suited for large-area applications. To demonstrate these capabilities, we covered the chest of a human-sized mannequin with the sensor array and also applied it to both the surface and end-effector of a UR5e robotic arm (**Fig. 7**).

We then subjected these setups to a series of extreme tests designed to evaluate both the sensor's spatial and temporal resolution under high-speed and high-impact conditions. For the mannequin, we simulated various forms of physical contact. First, we punched the sensor-covered area with boxing gloves to mimic sudden, forceful impacts. Next, we shot the mannequin using a Nerf gun to assess the sensor's ability to detect and localize fast, projectile-like forces. Finally, we used a yardstick to deliver slashing movements, testing the sensor's response to elongated, sweeping forces that vary in contact area and force distribution (**Video V7**).

For the UR5e, we conducted additional interaction tests to observe the sensor's durability and sensitivity to direct physical contact and pressure. We applied lateral force to the robot by pushing it and further tested impact resistance by striking it with the yardstick. To simulate a softer, continuous force application, we also programmed the robot to press down on a soft "brain" toy while capturing sensor readings (**Video V8**).

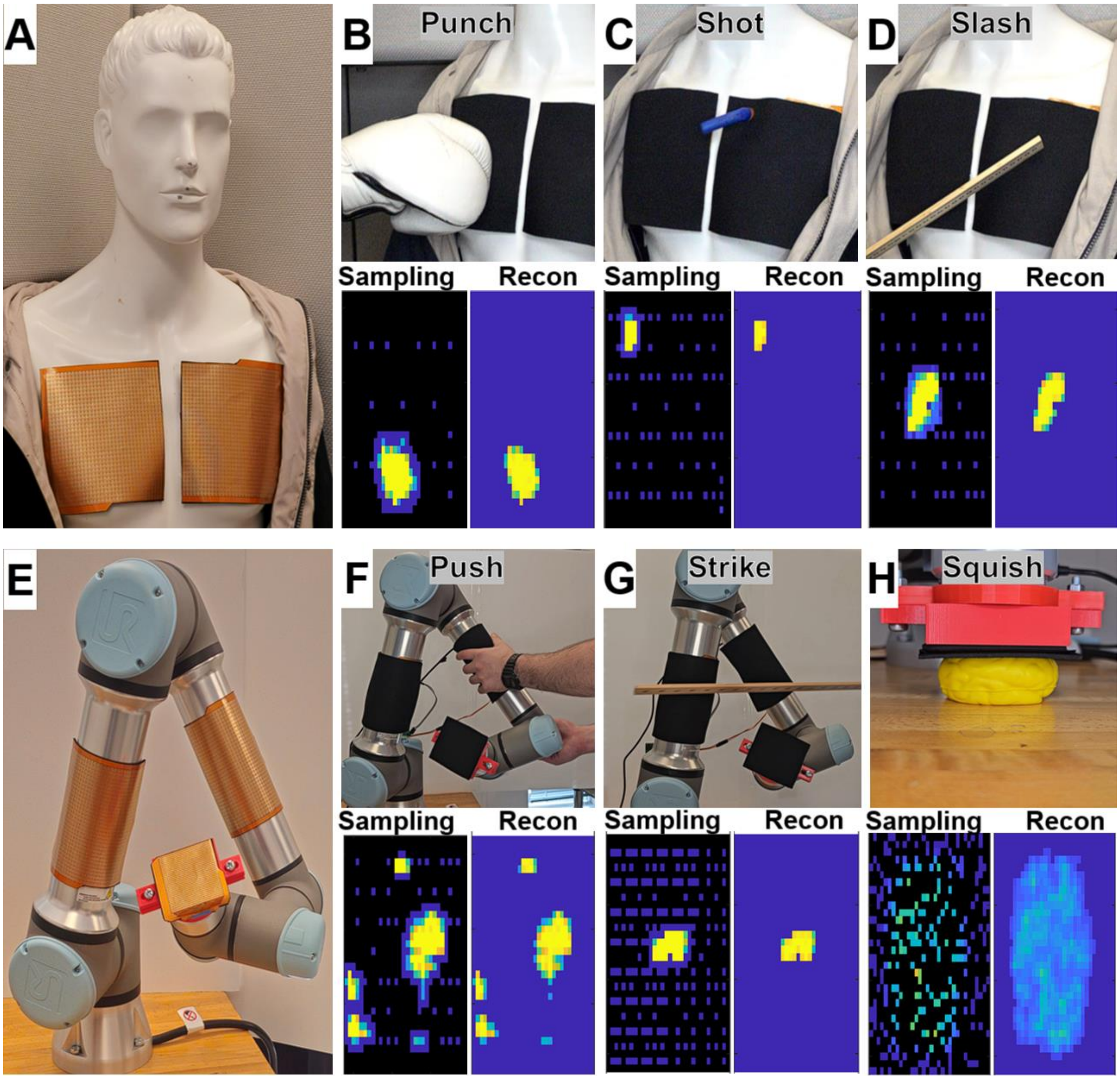

**Fig. 7. Large-Area Spatiotemporal Sensing. (A)** Mannequin covered with two sensor arrays mounted as chest-plates. **(B-D)** Image (top) and subsampling and reconstruction (bottom), of Punch, Shot, and Slash, respectively. **(E)** UR5e Robot covered with two sensor arrays on its arms, and a high-density sensor array mounted on its end effector. **(F-H)** Image (top) and subsampling and reconstruction (bottom), of Push, Strike, and Squish, respectively. Squish used the random sampling strategy, while the other tests used adaptive sampling. For all sampling and reconstruction plots, only the impacted sensor data is shown.

### *Wearable Applications for Full-Body Touch Sensing*

Because ACTS can be implemented in the firmware of a conventional tactile readout board, ACTS can connect and boost the spatiotemporal performance of any standard resistive tactile sensor matrix. To demonstrate this, and the potential of compressive tactile subsampling for enabling full-body, high-resolution touch sensing, we integrated our sensor into a variety of wearable formats, including a chest plate, a tactile sensing glove, a pressure sensing insole, a helmet, and a leg plate. These sensors were tested on a human participant, serving as an inspiration for future humanoid implementations (**Fig. 8**). The sensor on the arm (**Fig. 8B**) detected the rapid onset of a friendly arm tap, demonstrating its ability to sense localized impacts. Similarly, the tactile sensing glove worn on the hand (**Fig. 8C**) successfully identified the swift interaction of catching a ball, highlighting its responsiveness to high-speed events. The insole, placed inside a shoe (**Fig. 8D**), detected the rapid onset of a footstep, emphasizing its capability to monitor dynamic ground interactions. When mounted on a helmet worn on the head (**Fig. 8E**), the sensor detected the impact of a fast-moving projectile, showcasing its responsiveness to high-velocity forces. On the chest (**Fig. 8F**), the sensor registered the touch of a close and friendly interaction, demonstrating its sensitivity to softer, human-like contacts. Lastly, the sensor worn on the leg (**Fig. 8G**) effectively tracked the rhythmic motion of juggling a soccer ball, proving its ability to monitor continuous, repetitive movements. All sensor arrays contain 1024 sensing pixels. Demonstration video showing the spatiotemporal tactile signal from insole, glove, and wearable plates is shown in **Video V9**.

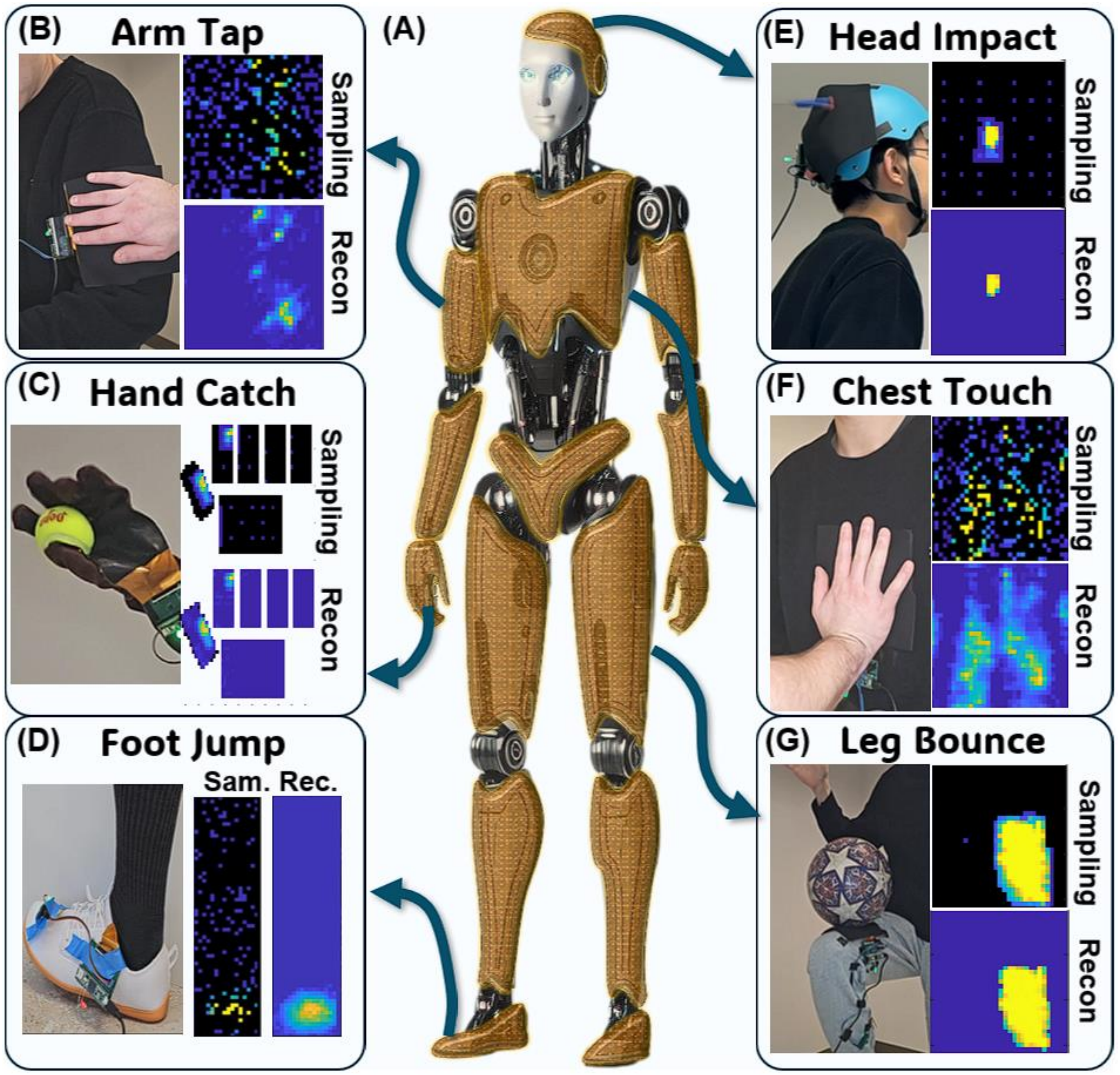

**Fig. 8. Wearable Applications for Full-Body Touch Sensing**. **(A)** Conceptual illustration of a humanoid robot covered in high-speed, high-resolution, full-body tactile skin enabled by compressive tactile subsampling. **(B)** Sensor worn on the arm detects the rapid onset of a friendly arm tap. **(C)** Sensor worn on the hand identifies the swift interaction of catching a ball. **(D)** Sensor worn on the foot, integrated into an insole, captures the dynamic onset of a footstep. **(E)** Sensor mounted on a helmet detects the impact of a fast-moving projectile. **(F)** Sensor worn on the chest registers a close, friendly touch. **(G)** Sensors worn on the leg monitor the rhythmic motion of juggling a soccer ball.

### *Real-time Embedded Compressive Tactile Reconstruction and Classification*

To further demonstrate the utility of compressive tactile sensing for real-time applications, the sparse recovery algorithm was ported to the MCU to consecutively run sensor measurement and tactile reconstruction and classification (**Fig. 9**). At small dictionary sizes (K=50) and low measurement levels (M=50), full sensor reconstruction (sparse recovery and dictionary multiplication) can be computed in less than 500 μsec. Therefore, sensor reconstruction (**Fig.**

**9B**) can occur sequentially to data acquisition and nominally yield a 720 FPS total frame rate (13x faster than the control of 55 FPS). Likewise, sensor classification (**Fig. 9A**) can occur sequentially to data acquisition and yield an 806 FPS total frame rate (15x faster). Sampling without reconstruction at M=50 would yield 1115 FPS. The relationship for real-time classification between measurement level, dictionary size, and frame rate is shown in **Fig. 9C**. Contour lines show the associated classification accuracies at different combinations of measurement level and dictionary size.

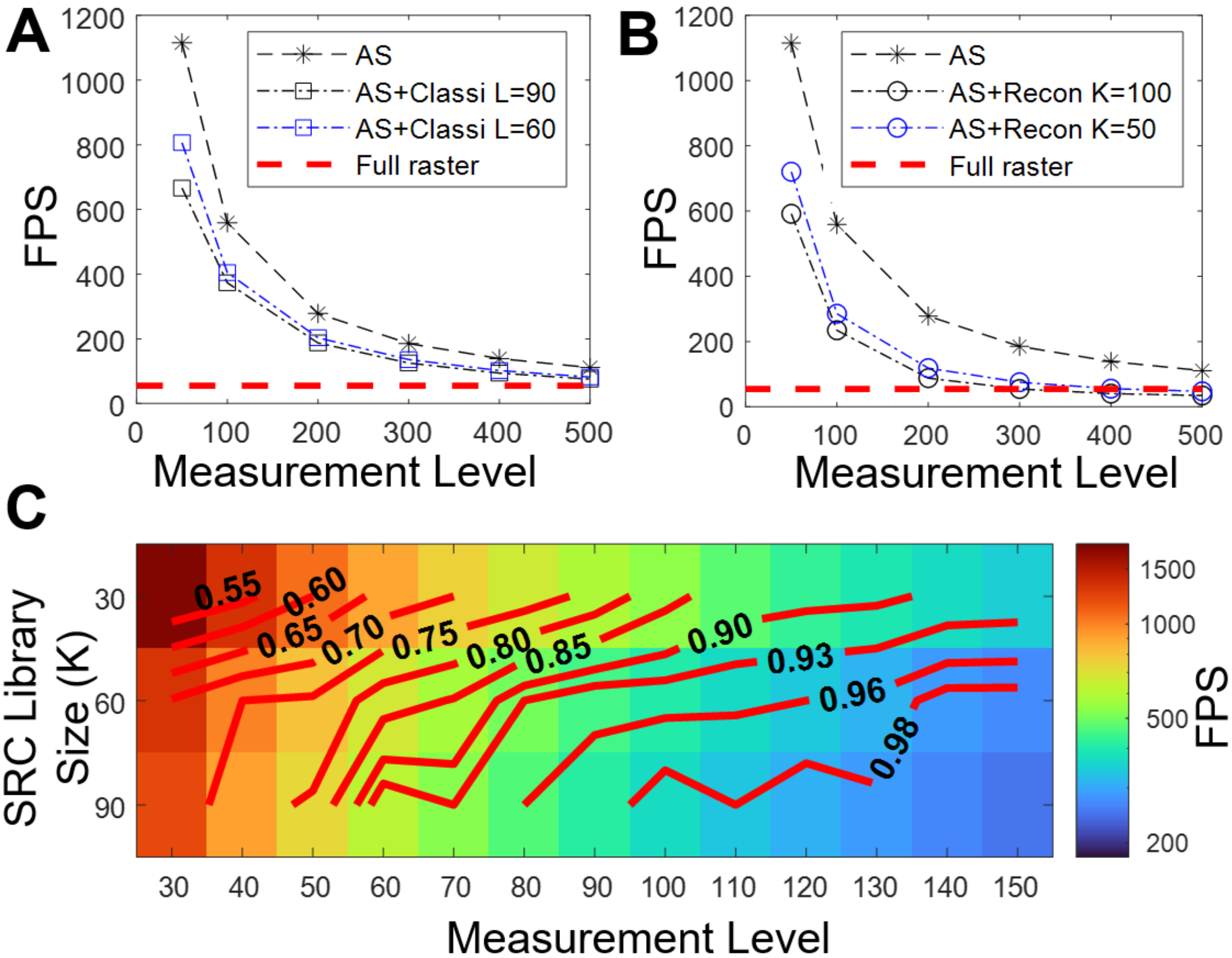

**Fig. 9. Real-time embedded reconstruction and classification. (A)** Frame rate during real-time classification at different dictionary sizes and measurement levels. L denotes the dictionary (library) size. **(B)** Frame rate during real-time reconstruction at different dictionary

### *Generalization to Unseen Objects and Contact Locations*

A key strength of ACTS is its ability to generalize to new, unseen tactile stimuli. This capability arises because the tactile dictionary captures fundamental local patterns of tactile interaction – such as edges, curves, and pressure gradients – rather than memorizing object-specific templates. As a result, ACTS can effectively interpret and reconstruct tactile imprints from objects of diverse sizes, shapes, and materials, even when those specific objects were not included in the dictionary learning process.

To rigorously demonstrate this generalization capability, we trained the dictionary using data from 20 distinct objects and evaluated ACTS on a held-out test set comprising 10 entirely novel objects. As shown in **Supplementary Figure S3-1**, the support accuracy versus measurement level *M* exhibits nearly identical trends for both the training and test sets across all sampling schemes. This consistency confirms that ACTS's learned dictionary encodes broadly applicable tactile primitives and enables robust compressed sensing on unseen stimuli.

To further illustrate adaptability, we also tested two additional objects ("Square" and "Two Cylinders") that were excluded from the dictionary and indented into the sensor at three new spatial positions (**Supplementary Fig. S3-2A–B**). The resulting measurements and reconstructions (**Fig. S3-2E–F**) show high support accuracy across all positions. Because the Square covers a larger contact area, random sampling provided better support estimates at low *M*, while the smaller Two Cylinders object was efficiently characterized by adaptive sampling. In both cases, reconstruction performance remained consistent with in-distribution examples, underscoring ACTS's robustness to variations in object geometry and placement.

### *Tactile-Based Closed-Loop Control and Balancing*

To demonstrate ACTS as an enabling technology for high-speed tactile feedback control, we implemented a robotic ball-balancing task on a UR5e arm, inspired by the dexterous object manipulation seen in Chinese walnut rolling. Crucially, the system relies only on tactile feedback from our sensor array to achieve fine motor control of a metallic ball.

This feedback control utilizes ACTS's tactile stream exceeding 1000 Hz for real-time dynamics estimation. The ball's position (derived from the array's center of pressure) and the desired trajectory served as inputs to a PID controller. The controller dynamically computed tilt angles for the UR5e's wrist joints, guiding the ball along predefined trajectories or maintaining balance at a fixed point.

We successfully demonstrated the system tracing complex patterns, including the abbreviation "NBI" representing our lab name (**Fig. 10, Video V11**), and balancing the ball near the center of the sensor array despite small disturbances (**Video V10)**. This application validates ACTS not only as a high-speed tactile readout solution, but also as a practical enabler of tactile-based feedback control for fine motor manipulation.

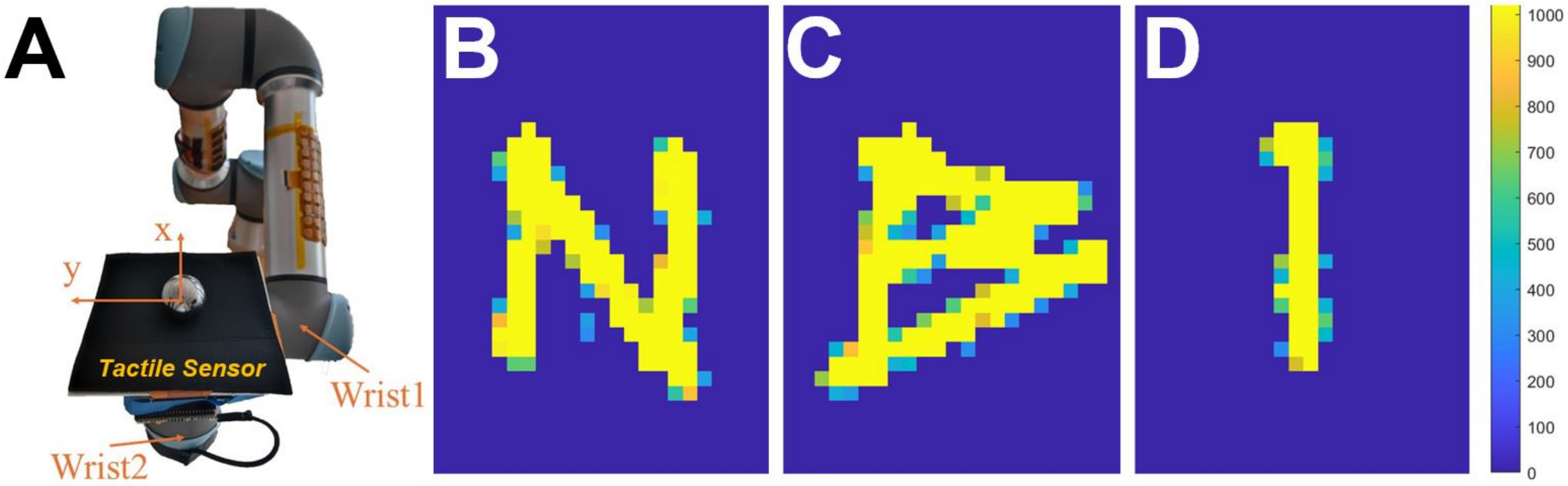

**Fig. 10. Tactile-based ball control and balance using compressive subsampling. (A)** UR5e robotic arm setup with tactile sensor array mounted on the end effector. **(B–D)** Cumulative pressure maps showing successful ball trajectory tracing of the letters "N", "B", and "I", respectively, using PID control and high-speed tactile feedback.

### *Whole-Hand Dexterous Tactile Reflex for Robotic Hand*

We integrated ACTS Skin with the LEAP robotic hand and executed tactile-only reflex behaviors that require coordinated, whole-hand sensing and control: spherical object grasping **(Fig. 11B**), hand-to-hand grasping (**Fig. 11C**), slip prevention (**Fig. 11D**), and sharp-object avoidance (**Fig. 11F**); representative trials are shown in **Video V12**. During spherical grasps, regional contact balancing stabilized the object at the palm center while preserving fingertip contact; for small objects, lateral comparisons induced a trapping rotation that increased contact uniformity. In hand-to-hand grasps, the controller redistributed normal forces across phalanges to maintain contact despite pose changes, without vision or external force sensing.

Under perturbations, the slip reflex detected tangential motion from successive tactile frames and increased grip effort until slip ceased, recovering from externally induced disturbances within one to two controller cycles. The sharp-avoidance reflex identified concentrated, tip-like features on the contact map and triggered immediate joint opening on the affected fingers, preventing further penetration while preserving neighboring contacts. Across tasks, the closed loop ran at 60 Hz, and actuation updates followed a discrete step rule with per-motor rate shaping. Effective hand speed and grip strength were governed by the base step size, motor scaling, and current limits.

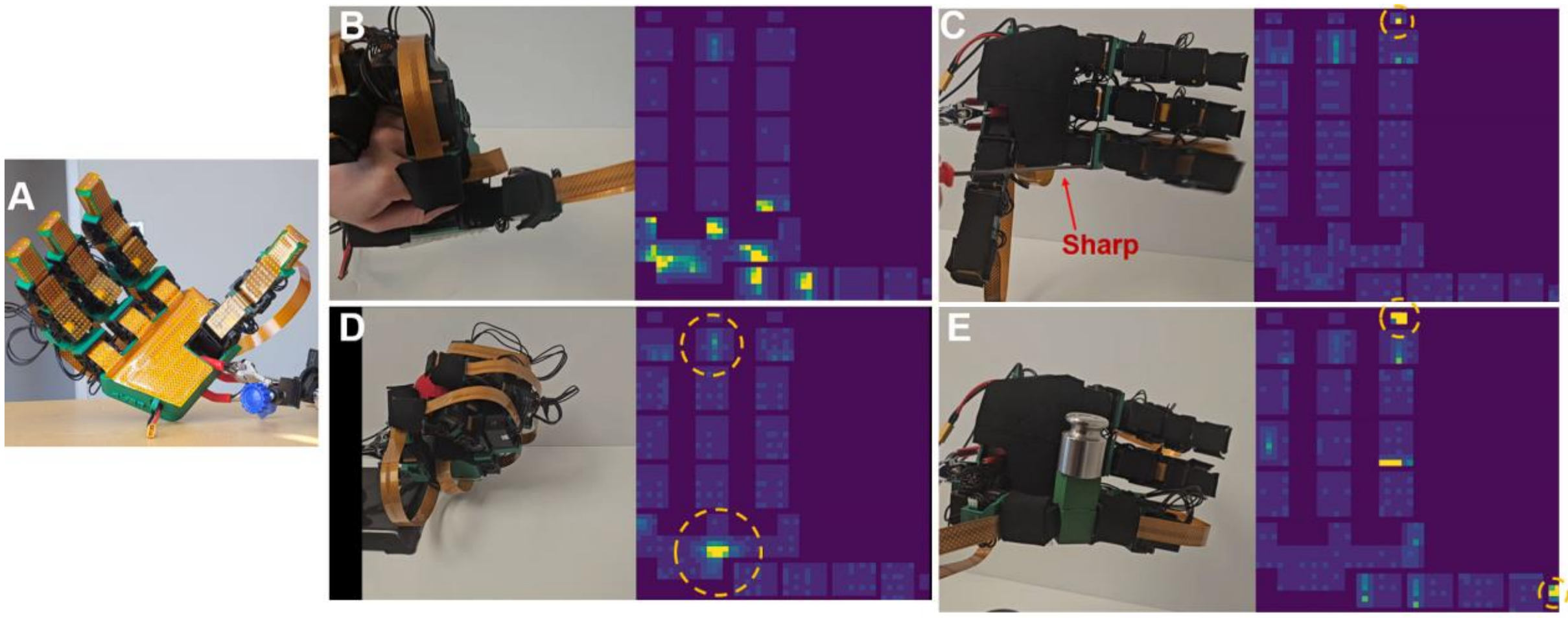

**Fig. 11. Tactile-only reflexive grasping in a dexterous robotic hand. (A)** Full-hand tactile sensorization of LEAP Hand (1024 sensors). **(B)** Stable grasp of a human hand via regional contact balancing. **(C)** Sharp-object avoidance detects a sharp tip and triggers an open-posture reflex. **(D)** Spherical grasp stabilized at the palm center through interior–exterior rebalancing and lateral reorientation. **(E)** Slip prevention under rapid loading on consecutive tactile frames to increase grip effort until slip ceases.

## Discussion

The compressive tactile subsampling approach presented in this work addresses a longstanding challenge in large-area tactile sensor arrays: achieving simultaneous high frame rates and high-resolution sensing without adding additional circuitry at the sensor level. By using a conventional readout circuit and deploying compressive subsampling through firmware, our approach can enhance tactile frame rates by 12X without compromising classification accuracy (99%, **Fig. 4F**), or 18X with slightly lower accuracy (90%). This is a significant innovation as researchers and engineers working in high-resolution tactile sensing can easily adopt our software technique to instantly boost the frame rate of their tactile sensor by >12X to achieve accurate dynamic sensing.

By boosting HD tactile frame rates to 1000 FPS, compressive tactile subsampling enables real-time detection of fast transient events such as deformable contact (**Fig. 6**) and sudden collisions/projectiles (**Fig. 7**). Temporal precision in these tasks can improve dynamic object manipulation tasks and also enhance safety during interactions with humans by providing rapid feedback to a robot about dynamic interactions.

Real-time reflex and reaction for human activities is often required to occur in as little as 150 ms (*51*), and human brains are known to process tactile sensory signals in less than 30 ms (*52*). Therefore, it is essential for artificial tactile sensing systems to rapidly identify and process transient contacts as well. Compressive tactile subsampling's ability to rapidly classify objects within the first 20 milliseconds of initial contact showcases the method's potential for manipulation tasks that require immediate feedback; for example, to perform human-like grasping. Additionally, the system's ability to track high-speed impacts highlights its usefulness in dynamic environments. This opens opportunities for applying compressive tactile subsampling to robotic systems with whole-body tactile sensors, where fast feedback is needed across large areas to ensure stable, adaptive interactions and human-level abilities.

Large-area coverage is a critical challenge in tactile sensor design for robots, yet it remains inadequately addressed by existing solutions. However, ACTS enables the transformation of standard passive tactile arrays into high-speed systems, making them highly adaptable for large-area applications. Unlike current neuromorphic approaches, which often compromise on robustness or sensor density, ACTS maintains both, allowing seamless integration across entire robots or human bodies. This capability is demonstrated through its successful implementation on diverse platforms, including a UR5e robotic arm, a mannequin, and a human body. These examples highlight ACTS's potential to deliver reliable, dense, and high-speed tactile sensing over extensive and complex surfaces without sacrificing performance, paving the way for scalable and practical touch-based solutions in robotics and wearable technologies.

Beyond perception and classification, ACTS also enables closed-loop control tasks using tactile feedback alone. To demonstrate this, we implemented a tactile-based ball balancing system where a UR5e robot steers a metallic ball across a tactile sensor array to trace complex trajectories. By detecting the ball's position in real-time via compressive subsampling and adjusting the tilt of the array accordingly, the robot successfully executed trajectory tracking and balancing behaviors. This highlights the utility of ACTS not only for high-speed perception, but also for tactile servoing and motor control in dynamic environments. These results extend the scope of ACTS from reactive sensing to full tactile-based closed-loop control, unlocking new possibilities for dexterous manipulation, balance maintenance, and shape-following in robotics. Extending to a dexterous end-effector, we instrumented the full LEAP hand with a 1024-element ACTS Skin and ran tactile-only whole-hand reflexes at 60 Hz. The controller stabilized spherical and hand-to-hand grasps, countered externally induced slip by increasing grip effort, and executed sharp-object avoidance via rapid joint opening; all without vision or external force sensing (Fig. 11; Video V12). These results demonstrate that the same compressed-sensing pipeline that enables high-speed perception also supports closed-loop manipulation on multi-DOF hands. Practically, this suggests a path from high-rate tactile perception to scalable, reflexive dexterity on fully sensorized robotic hands.

In this work random, adaptive, and uniform subsampling were considered as three separate sampling schemes, however combinations or hybrid approaches can be more efficient in many cases. For example, since adaptive sampling performs best for small objects and random sampling performs best for large objects, a hybrid approach that used adaptive pixel selection followed by random subsampling could prove more efficient for medium sized objects. Additionally, the three sensing methods tested were not optimized based on the tactile dictionary. This is another area for improvement where the sensing scheme can be enhanced based on the dictionary (*53*).

A key feature of our system is its ability to run efficiently and in real-time on an Arduino-like microcontroller. By developing on a device in the Arduino ecosystem, real-time tactile reconstruction and classification at high frame rates is made highly accessible to teams without embedded systems expertise. On the other hand, bolstering the choice of MCU, for example by choosing one with more processing cores, could lead to even lower latencies as signal reconstruction can occur in parallel to measurement acquisition.

## Materials and Methods

### *Tactile Dictionary Learning*

Tactile dictionary learning was performed using K-SVD (*54*). The 30 objects in the object library were pressed into the sensor array using the robotic arm and frames above a threshold value were extracted for each object. These frames were divided into random 8x8 patches selected across the tactile frame. Repetitive and redundant frames with high coherence calculated as: $\mu(\mathbf{A}) = \max_{1 \le i \ne j \le N} \left| \langle \mathbf{a}_i, \mathbf{a}_j \rangle \right|$ were removed. Frames with few active pixels were also removed. The training dataset consists of 14,907 patches of size 8x8, each with more than 15 non-zero pixels, selected from 100 full-raster tactile images of 10 objects. The K-SVD algorithm is configured with 10 iterations and a sparsity level of 13. The dictionary size for each patch is 1,000. For the real-time tactile reconstruction experiment, the training dataset comprises 1,492 patches, with dictionary sizes for each patch set at 50, 100, 200, 300, 400, and 500 for the respective experiments.

Main experiments (Figs. 4–6) train the dictionary on frames drawn from the object library (see above) and evaluate on the same library under held-out trials. For generalization experiments (Fig. S3-1/S3-2), we train a separate dictionary on 20 objects and evaluate on 10 disjoint, novel objects and on two additional objects (‘Square’, ‘Two Cylinders’) pressed at three new positions each. Dictionary training, patch sampling, and preprocessing are identical across regimes.

### *Sparse Recovery and Tactile Reconstruction*

Sparse recovery was implemented using a modified “FastOMP” algorithm that is hardware friendly (*49*). The sparsity of sparse recovery is set to 25% of the measurement level for a given reconstruction. Tactile frames were reconstructed in 8x8 patches with an overlap of 4 pixels. Reconstructions were averaged in the overlapping regions. The interpolation control cases were implemented using the MATLAB “scatteredInterpolant” function using linear interpolation and extrapolation.

### *Sparsity Quantification*

We quantify active-taxel fraction on full-raster frames from the object library to characterize dataset sparsity. During recovery, we set the target sparsity to S = 0.25M, consistent with compressed-sensing practice and our empirical analysis. **Fig. S4** shows diminishing returns in reconstruction accuracy beyond ~25% sparsity, supporting this operating point.

### *Design of Tactile Sensor Arrays*

All tactile sensor arrays were developed with flexible printed circuit boards (fPCB). The sensor arrays are arranged as a 32x32 matrix, with varying resolutions (**Supplemental. S10**). The sensors use a single layer design with row and column electrodes being interdigitated. The pressure sensing insole uses an overlapping 2-layer design. A piezoresistive material (Velostat, 3M) is applied above the sensor array to bridge the gap between the electrodes. An adhesive foam (NATGAI Sponge Neoprene with Adhesive Foam Rubber Sheet) is applied above the piezoresistive material to protect the sensor and distribute the force. The fPCB is designed with an FPC connector to mate with a Molex connector on the readout board.

***Hardware Design***

The readout board is based on the Teensy 4.1 MCU (ARM Cortex-M7, NXP iMXRT1062). The MCU has 32 digital connections to the rows of the tactile sensor array and controls a 32:1 analog multiplexer (ADG732BSUZ) to switch different columns of the array to a readout circuit. The readout circuit is a summing amplifier based on the (TLV9362). An additional inverting amplifier is added after the summing amplifier to ensure the output voltage remains within the readable range of the ADC inside the MCU. The readout board features an Easy-On FFC/FPC Molex connector with 70 circuits (5051107091) for the sensor array to attach to. The use of the Teensy 4.1 allows the entire system to be programmable in the Arduino IDE.

***Robotic Interaction with Daily Objects***

The Universal Robotic UR5e cobot was used to perform repetitive indentation interactions between the sensor array and a library of 30 daily and 3D printed objects. 3D printed objects were fastened to the end effector of the robot using bolts, and daily objects were attached to the robot using a claw gripper. The robot repeatedly moved the object into the sensor at a repetitive spot and speed for each object. The object library included: 3D printed objects {"2 cylinders", "3 lines", "diagonal line", "3 rings", "triangle", "circle", "square", "X"} and daily objects {"computer mouse", "art scissors", "pliers", "wristwatch", "fidget spinner", "rubber ball", "hex bar", "AAA battery", "wrench", "bolt and nut", "spoon", "soft brain", "tennis ball", "glass mug", "multitool", "cologne", "glue bottle", "9V battery", "eraser", "power adapter", "screwdriver", "tweezers"}.

***Subsampling Methods***

Tactile subsampling was performed using three different subsampling schemes, including Random, Uniform, and Adaptive.

Uniform subsampling selects M pixels of the tactile array to sample using a constant 'row' and 'column' offset. Sequential frame samples rotate the uniform subsampled pixels to ensure that all pixels in the sensor array will eventually be sampled (**Fig. S13**).

Random subsampling randomly selects M pixels of the tactile array to sample for each frame.

The pseudocode for Adaptive subsampling for one frame is shown in **Supplemental Algorithm 1**. Adaptive subsampling is an adaptive sampling scheme that modifies the order of pixels to be examined dynamically in response to their measurements. Our adaptive method works by splitting vertically or horizontally the search space of the tactile array into 2 equal parts and sampling the center pixel for each of the spaces. If a sampled pixel $p$ records a measurement above a threshold, neighboring pixels are sampled ($\mathrm{NeighborList}(p)$), with an order shown in **Fig. S15A**. This adaptive process will continue recursively until the force of all surrounding pixels is lower than a threshold ($NSThr$ in Algorithm 1). Each space is split further, and pixels are sampled with patterns as described above. This process repeats until M measurements are taken. During this process, there is no repetitive measured pixel for each tactile frame. Here, the search space is assumed to be square, that is N is a square number. Note that in Algorithm 1, $\mathrm{ForceAt}(p)$ means examining force at position $p$. The full pattern of Adaptive sampling on a 32x32 square tactile sensor can be visualized in **Fig. S15B**, and an example pattern on an 15x15 sensor ($N = 225$) with $M = 55$ can be seen in **Fig. S15C**.

### *SRC Classification*

Our SRC classification uses a conventional SRC approach (*55*). Key high-pressure frames of the objects to classify were saved into a library matrix. Following subsampling, sparse recovery is computed to find the representative atoms in the library matrix. A signal is reconstructed based on the sparse representation and is subtracted from each of the library examples. The class of the experimental object is assigned as the library example with the smallest residual after subtraction.

### *Real-time Embedded Sparse Recovery*

Real-time sparse recovery was implemented on the Teensy 4.1 using the modified FastOMP algorithm. To ensure real-time recovery, a few optimizations were performed that boosted computation speed. For example, these include saving the dictionary matrix once and saving a list of the positions of the patches to reconstruct. This saves unnecessary copying of the dictionary matrix. Additionally, for the FastOMP algorithm rather than representing the long sparse vector with an array of mostly zeros, the sparse vector was saved as 2 shorter arrays: one with the list of non-zero values and another with their index in the sparse vector.

### *High-speed projectile and deformable interactions*

A tennis ball was dropped onto the sensor from table height and the measurement level of compressive tactile sensing was varied. The tennis ball would bounce off the sensor, contacting the sensor for approximately 8 ms. For the deformable object experiment, the deformable objects were bounced onto the sensor from a 10 cm height.

### *Foam-bullet Tracking*

A Nerf N-Strike Elite Disruptor was used to fire foam bullets at the sensor from 2 meters away. A slow-motion camera (Sony Cyber-Shot DSC-RX100 V) was used to record the interaction at 960 FPS.

### *Ricochet Angle Estimation*

A rubber ball was rolled off a metal ramp to bounce off the sensor at a specific angle. The angle of the ramp was adjusted for 8 different angles, (0°, 45°, 90°, 135°, 180°, -135°, -90°, -45°). The center of pressure (COP) was computed for each frame during the bounce portion of the ricochet. The progression of the COP was tracked, and the total COP displacement was the estimated ricochet angle. The slow-motion camera captured the ricochet from a top view.

### *Effect of other dictionary choice*

To compare the learned tactile dictionary to a standard dictionary, we repeated the experiment in "Rapid Tactile Reconstruction and Classification of Daily Objects" with an overcomplete DCT dictionary and an overcomplete wavelet dictionary (Haar). The overcomplete DCT was generated using the KSVD-box 13 library (*56*), and overcomplete Haar was generated by changing the shifts and maximum level of the wavelet matrix generated with the WavMat function (*57*).

***Ball Control and Trajectory Tracking***

To demonstrate the utility of high-speed tactile feedback for closed-loop control, we implemented a tactile-based ball balancing system using a UR5e robotic arm. A metallic ball was placed on a planar tactile sensor array (32×32) mounted on the UR5e end effector. The arm controlled the plate's orientation using its two wrist joints to induce tilt along two perpendicular axes, thereby generating in-plane accelerations through gravity. By adjusting these tilt angles, the system steered the ball's motion across the surface in real time. Full description of control setup is shown in Supplemental S16.

The ball's estimated planar position (determined at each timestep from the highest-pressure taxel) and the desired trajectory were used as inputs to a PID controller that computed the necessary tilt angles. Implemented in ROS2 (Humble), the system ensured reliable synchronization between the 1000 Hz tactile sensor and the 30 Hz control loop.

We evaluated this tilt-controlled platform on static and dynamic tasks. The system maintained the ball balanced at the array center despite perturbations and guided it accurately along trajectories tracing the letters "N", "B", and "I". Cumulative pressure maps for these trajectories (Fig. 10) validate closed-loop control using solely tactile feedback.

L***EAP Hand Reflex Controller.***

A 1024-taxel ACTS Skin was integrated on the LEAP hand. Tactile frames were streamed to a controller at 60 Hz. For each frame, we compute regional contact maps (fingertips, phalanges, palm), total normal force, and frame-to-frame tangential change.

Reflexes were implemented as follows: (i) Slip prevention: detect tangential motion from successive frames; increase grip command until slip ceases. (ii) Sharp-object avoidance: detect concentrated tip-like contacts (local peak/area ratio exceeding a preset threshold); trigger rapid opening of the affected joints while preserving neighboring contacts. (iii) Spherical grasp balancing: compare interior vs. exterior palm regions and fingertips; redistribute normal force and induce lateral re-orientation for small objects to centralize contact. Control logic ran without vision or external force sensing (Video V12).

**Supplementary Materials**

S1. Table of large-area tactile sensor arrays without integrated sensor electronics with number of sensors, sampling rate, and the total sensor rate.

S2. Figure of merit plot showing total sensor rate vs number of measurements per frame

S3. Generalizability of Compressive Subsampling to New Objects

S4. Effect of Sparsity Level on Reconstruction Accuracy

S5. Learned Tactile Dictionary

S6. Object Size vs Accuracy

S7. Confusion matrices for classification of objects

S8. Compared with other dictionaries (wavelet and DCT)

S9. Force response of taxel

S10. Circuit board and tactile sensor layout

S11. Component list

S12. Readout board schematic

S13. Picture of the 30 objects

S14. Frame-wise shifting pattern of uniform subsampling method

S15. Pattern of adaptive sampling with neighboring search

S16. Robot tactile-based control for ball balancing

V1. Video of tennis ball impacting sensor and spatiotemporal tactile reconstruction

V2. Video of ricochetting balls and estimation of ricochet angle using tactile data

V3. Video of foam-bullet detection fired by NERF gun

V4. Video of dictionary learning process

V5. Video of changing from low-density to high-density sensor

V6. Video of deformable object collision

V7. Video of punching, slashing, and shooting mannequin

V8. Video of UR5e Cobot squeezing foam brain

V9. Video of full body wearable sensors, glove, insole, arm/leg/chest/plates

V10. Video tactile-based robotic ball balancing

V11. Video of tactile-guided letter tracing with robotic ball control

V12. Video of tactile-only reflexes on fully sensorized LEAP hand\

Videos can be accessed at this link: ACTS Videos

Data, code, and design files are all found in our GitHub repository:
https://github.com/aslepyan/CompressiveTactileSubsampling

**References:**

1. J. A. Fishel, B. Matulevich, K. A. Muller, G. M. Berke, "The (Sensorized) Hand is Quicker than the Eye: Restoring Grasping Speed and Confidence for Amputees with Tactile Reflexes" in *2019 International Conference on Robotics and Automation (ICRA)* (IEEE, Montreal, QC, Canada, 2019; https://ieeexplore.ieee.org/document/8793643/), pp. 5097–5102.
2. E. Luberto, Y. Wu, G. Santaera, M. Gabiccini, A. Bicchi, Enhancing Adaptive Grasping Through a Simple Sensor-Based Reflex Mechanism. *IEEE Robot. Autom. Lett.* **2**, 1664–1671 (2017).
3. H. Deng, Y. Zhang, X.-G. Duan, Wavelet Transformation-Based Fuzzy Reflex Control for Prosthetic Hands to Prevent Slip. *IEEE Trans. Ind. Electron.* **64**, 3718–3726 (2017).
4. T. Dahl, A. Palmer, Touch-triggered protective reflexes for safer robots. (2010).
5. Z. Kappassov, J.-A. Corrales, V. Perdereau, Tactile sensing in dexterous robot hands – Review. *Robotics and Autonomous Systems* **74**, 195–220 (2015).
6. H. Yousef, M. Boukallel, K. Althoefer, Tactile sensing for dexterous in-hand manipulation in robotics – A review. *Sensors and Actuators A: Physical* **167**, 171–187 (2011).
7. F. R. Hogan, J. Ballester, S. Dong, A. Rodriguez, "Tactile Dexterity: Manipulation Primitives with Tactile Feedback" in *2020 IEEE International Conference on Robotics and Automation (ICRA)* (IEEE, Paris, France, 2020; https://ieeexplore.ieee.org/document/9196976/), pp. 8863–8869.
8. L. L. Bologna, J. Pinoteau, J.-B. Passot, J. A. Garrido, J. Vogel, E. R. Vidal, A. Arleo, A closed-loop neurobotic system for fine touch sensing. *J. Neural Eng.* **10**, 046019 (2013).
9. H. Oh, G.-C. Yi, M. Yip, S. A. Dayeh, Scalable tactile sensor arrays on flexible substrates with high spatiotemporal resolution enabling slip and grip for closed-loop robotics. *Sci. Adv.* **6**, eabd7795 (2020).

10. W. Yang, M. Hon, H. Yao, B. C. K. Tee, *An Atlas for Large-Area Electronic Skins: From Materials to Systems Design* (Cambridge University Press, ed. 1, 2020; https://www.cambridge.org/core/product/identifier/9781108782395/type/element).
11. S. Luo, J. Bimbo, R. Dahiya, H. Liu, Robotic tactile perception of object properties: A review. *Mechatronics* **48**, 54–67 (2017).
12. M. Soni, R. Dahiya, Soft eSkin: distributed touch sensing with harmonized energy and computing. *Phil. Trans. R. Soc. A.* **378**, 20190156 (2020).
13. W. W. Lee, Y. J. Tan, H. Yao, S. Li, H. H. See, M. Hon, K. A. Ng, B. Xiong, J. S. Ho, B. C. K. Tee, A neuro-inspired artificial peripheral nervous system for scalable electronic skins. *Sci. Robot.* **4**, eaax2198 (2019).
14. W. W. Lee, J. Cabibihan, N. V. Thakor, "Bio-mimetic strategies for tactile sensing" in *2013 IEEE SENSORS* (IEEE, Baltimore, MD, USA, 2013; http://ieeexplore.ieee.org/document/6688260/), pp. 1–4.
15. S. Sundaram, P. Kellnhofer, Y. Li, J.-Y. Zhu, A. Torralba, W. Matusik, Learning the signatures of the human grasp using a scalable tactile glove. *Nature* **569**, 698–702 (2019).
16. Y. Luo, Y. Li, P. Sharma, W. Shou, K. Wu, M. Foshey, B. Li, T. Palacios, A. Torralba, W. Matusik, Learning human–environment interactions using conformal tactile textiles. *Nat Electron* **4**, 193–201 (2021).
17. W. W. Lee, S. L. Kukreja, N. V. Thakor, Discrimination of Dynamic Tactile Contact by Temporally Precise Event Sensing in Spiking Neuromorphic Networks. *Front. Neurosci.* **11** (2017).
18. O. Oballe-Peinado, J. A. Hidalgo-Lopez, J. A. Sanchez-Duran, J. Castellanos-Ramos, F. Vidal-Verdu, "Architecture of a tactile sensor suite for artificial hands based on FPGAs" in *2012 4th IEEE RAS & EMBS International Conference on Biomedical Robotics and Biomechatronics (BioRob)* (IEEE, Rome, Italy, 2012; http://ieeexplore.ieee.org/document/6290746/), pp. 112–117.
19. R. S. Dahiya, G. Metta, M. Valle, G. Sandini, Tactile Sensing – From Humans to Humanoids. *IEEE Trans. Robot.* **26**, 1–20 (2010).
20. W. W. Lee, S. L. Kukreja, N. V. Thakor, "A kilohertz kilotaxel tactile sensor array for investigating spatiotemporal features in neuromorphic touch" in *2015 IEEE Biomedical Circuits and Systems Conference (BioCAS)* (IEEE, Atlanta, GA, USA, 2015; http://ieeexplore.ieee.org/document/7348412/), pp. 1–4.
21. Ó. Oballe-Peinado, J. A. Hidalgo-López, J. Castellanos-Ramos, J. A. Sánchez-Durán, R. Navas-González, J. Herrán, F. Vidal-Verdú, FPGA-Based Tactile Sensor Suite Electronics for Real-Time Embedded Processing. *IEEE Transactions on Industrial Electronics* **64**, 9657–9665 (2017).
22. F. Bergner, P. Mittendorfer, E. Dean-Leon, G. Cheng, "Event-based signaling for reducing required data rates and processing power in a large-scale artificial robotic skin" in *2015 IEEE/RSJ International Conference on Intelligent Robots and Systems (IROS)* (IEEE, Hamburg, Germany, 2015; http://ieeexplore.ieee.org/document/7353660/), pp. 2124–2129.
23. C. Bartolozzi, P. M. Ros, F. Diotalevi, N. Jamali, L. Natale, M. Crepaldi, D. Demarchi, "Event-driven encoding of off-the-shelf tactile sensors for compression and latency optimisation for robotic skin" in *2017 IEEE/RSJ International Conference on Intelligent Robots and Systems (IROS)* (IEEE, Vancouver, BC, 2017; http://ieeexplore.ieee.org/document/8202153/), pp. 166–173.
24. Z. Long, W. Lin, P. Li, B. Wang, Q. Pan, X. Yang, W. W. Lee, H. S.-H. Chung, Z. Yang, One-wire reconfigurable and damage-tolerant sensor matrix inspired by the auditory tonotopy. *Sci. Adv.* **9**, eadi6633 (2023).
25. M. F. Duarte, M. A. Davenport, D. Takhar, J. N. Laska, T. Sun, K. F. Kelly, R. G. Baraniuk, Single-pixel imaging via compressive sampling. *IEEE Signal Process. Mag.* **25**, 83–91 (2008).
26. A. C. Sankaranarayanan, C. Studer, R. G. Baraniuk, "CS-MUVI: Video compressive sensing for spatial-multiplexing cameras" in *2012 IEEE International Conference on Computational Photography (ICCP)* (IEEE, Seattle, WA, USA, 2012; http://ieeexplore.ieee.org/document/6215212/), pp. 1–10.
27. R. F. Marcia, Compressed sensing for practical optical imaging systems: a tutorial. *Opt. Eng* **50**, 072601 (2011).
28. C. Quinsac, A. Basarab, J.-M. Girault, D. Kouame, "Compressed sensing of ultrasound images: Sampling of spatial and frequency domains" in *2010 IEEE Workshop On Signal Processing Systems* (IEEE, San Francisco, CA, USA, 2010; http://ieeexplore.ieee.org/document/5624793/), pp. 231–236.
29. R. Van Sloun, A. Pandharipande, M. Mischi, L. Demi, Compressed Sensing for Ultrasound Computed Tomography. *IEEE Trans. Biomed. Eng.* **62**, 1660–1664 (2015).
30. M. F. Schiffner, G. Schmitz, "Fast image acquisition in pulse-echo ultrasound imaging using compressed sensing" in *2012 IEEE International Ultrasonics Symposium* (IEEE, Dresden, Germany, 2012; http://ieeexplore.ieee.org/document/6561995/), pp. 1944–1947.
31. J. Zhang, S. Mitra, Y. Suo, A. Cheng, T. Xiong, F. Michon, M. Welkenhuysen, F. Kloosterman, P. S. Chin, S. Hsiao, T. D. Tran, F. Yazicioglu, R. Etienne-Cummings, A closed-loop compressive-sensing-based neural recording system. *J. Neural Eng.* **12**, 036005 (2015).
32. J. Zhang, Y. Suo, S. Mitra, S. Chin, S. Hsiao, R. F. Yazicioglu, T. D. Tran, R. Etienne-Cummings, An efficient and compact compressed sensing microsystem for implantable neural recordings. *IEEE Trans. Biomed. Circuits Syst.* **8**, 485–496 (2014).
33. Y. Suo, J. Zhang, T. Xiong, P. S. Chin, R. Etienne-Cummings, T. D. Tran, Energy-Efficient Multi-Mode Compressed Sensing System for Implantable Neural Recordings. *IEEE Trans. Biomed. Circuits Syst.* **8**, 0–0 (2014).
34. M. Shoaran, M. M. Lopez, V. S. R. Pasupureddi, Y. Leblebici, A. Schmid, "A low-power area-efficient compressive sensing approach for multi-channel neural recording" in *2013 IEEE International Symposium on Circuits and Systems (ISCAS2013)* (IEEE, Beijing, 2013; http://ieeexplore.ieee.org/document/6572310/), pp. 2191–2194.

35. W. L. Chan, M. L. Moravec, R. G. Baraniuk, D. M. Mittleman, Terahertz imaging with compressed sensing and phase retrieval. *Opt. Lett.* **33**, 974 (2008).
36. P. K. Baheti, H. Garudadri, "An Ultra Low Power Pulse Oximeter Sensor Based on Compressed Sensing" in *2009 Sixth International Workshop on Wearable and Implantable Body Sensor Networks* (IEEE, Berkeley, CA, 2009; http://ieeexplore.ieee.org/document/5226900/), pp. 144–148.
37. B. Hollis, S. Patterson, J. Trinkle, "Adaptive basis selection for compressed sensing in robotic tactile skins" in *2017 IEEE Global Conference on Signal and Information Processing (GlobalSIP)* (IEEE, Montreal, QC, 2017; http://ieeexplore.ieee.org/document/8309168/), pp. 1285–1289.
38. B. Hollis, S. Patterson, J. Trinkle, Compressed Learning for Tactile Object Recognition. *IEEE Robot. Autom. Lett.* **3**, 1616–1623 (2018).
39. B. Hollis, S. Patterson, J. Trinkle, "Compressed sensing for tactile skins" in *2016 IEEE International Conference on Robotics and Automation (ICRA)* (IEEE, Stockholm, Sweden, 2016; http://ieeexplore.ieee.org/document/7487128/), pp. 150–157.
40. B. Hollis, "Compressed Sensing for Scalable Robotic Tactile Skins," (2018).
41. L. Shao, T. Lei, T.-C. Huang, Z. Bao, K.-T. Cheng, "Robust Design of Large Area Flexible Electronics via Compressed Sensing" in *2020 57th ACM/IEEE Design Automation Conference (DAC)* (IEEE, San Francisco, CA, USA, 2020; https://ieeexplore.ieee.org/document/9218570/), pp. 1–6.
42. L. E. Aygun, P. Kumar, Z. Zheng, T.-S. Chen, S. Wagner, J. C. Sturm, N. Verma, Hybrid LAE-CMOS Force-Sensing System Employing TFT-Based Compressed Sensing for Scalability of Tactile Sensing Skins. *IEEE Trans. Biomed. Circuits Syst.* **13**, 1264–1276 (2019).
43. L. E. Aygun, P. Kumar, Z. Zheng, T.-S. Chen, S. Wagner, J. C. Sturm, N. Verma, "17.3 Hybrid System for Efficient LAE-CMOS Interfacing in Large-Scale Tactile-Sensing Skins via TFT-Based Compressed Sensing" in *2019 IEEE International Solid- State Circuits Conference - (ISSCC)* (IEEE, San Francisco, CA, USA, 2019; https://ieeexplore.ieee.org/document/8662442/), pp. 280–282.
44. W. W. Lee, S. L. Kukreja, N. V. Thakor, Discrimination of Dynamic Tactile Contact by Temporally Precise Event Sensing in Spiking Neuromorphic Networks. *Front. Neurosci.* **11** (2017).
45. A. Slepyan, M. Zakariaie, T. Tran, N. Thakor, Wavelet Transforms Significantly Sparsify and Compress Tactile Interactions. *Sensors* **24**, 4243 (2024).
46. W. Fukui, F. Kobayashi, F. Kojima, H. Nakamoto, N. Imamura, T. Maeda, H. Shirasawa, High-Speed Tactile Sensing for Array-Type Tactile Sensor and Object Manipulation Based on Tactile Information. *Journal of Robotics* **2011**, 1–9 (2011).
47. M. A. Davenport, M. F. Duarte, Y. C. Eldar, G. Kutyniok, "Introduction to compressed sensing" in *Compressed Sensing*, Y. C. Eldar, G. Kutyniok, Eds. (Cambridge University Press, ed. 1, 2012; https://www.cambridge.org/core/product/identifier/CBO9780511794308A008/type/book_part), pp. 1–64.
48. E. J. Candes, M. B. Wakin, An Introduction To Compressive Sampling. *IEEE Signal Process. Mag.* **25**, 21–30 (2008).
49. X. Ge, F. Yang, H. Zhu, X. Zeng, D. Zhou, An Efficient FPGA Implementation of Orthogonal Matching Pursuit With Square-Root-Free QR Decomposition. *IEEE Trans. VLSI Syst.* **27**, 611–623 (2019).
50. R. Yarahmadi, A. Safarpour, R. Lotfi, An Improved-Accuracy Approach for Readout of Large-Array Resistive Sensors. *IEEE Sensors J.* **16**, 210–215 (2016).
51. E. Tønnessen, T. Haugen, S. A. I. Shalfawi, Reaction Time Aspects of Elite Sprinters in Athletic World Championships. *Journal of Strength and Conditioning Research* **27**, 885–892 (2013).
52. T. Callier, A. K. Suresh, S. J. Bensmaia, Neural Coding of Contact Events in Somatosensory Cortex. *Cerebral Cortex* **29**, 4613–4627 (2019).
53. M. Elad, Optimized Projections for Compressed Sensing. *IEEE Trans. Signal Process.* **55**, 5695–5702 (2007).
54. M. Aharon, M. Elad, A. Bruckstein, K-SVD: An algorithm for designing overcomplete dictionaries for sparse representation. *IEEE Transactions on Signal Processing* **54**, 4311–4322 (2006).
55. J. Wright, A. Y. Yang, A. Ganesh, S. S. Sastry, Yi Ma, Robust Face Recognition via Sparse Representation. *IEEE Trans. Pattern Anal. Mach. Intell.* **31**, 210–227 (2009).
56. R. Rubinstein, M. Zibulevsky, M. Elad, Efficient implementation of the K-SVD algorithm using batch orthogonal matching pursuit.
57. B. Vidakovic, WavMat; https://gtwavelet.bme.gatech.edu/.
58. T. Mouri, H. Kawasaki, K. Yoshikawa, J. Takai, S. Ito, Anthropomorphic Robot Hand "Gifu Hand III" and Real Time Control System. *Robomech* **2002**, 112 (2002).
59. Pressure Profile Systems, Capacitive Tactile Sensors in Robotics (2016). https://pressureprofile.com/tactarray/conformable-tactarray.
60. TekScan, I-Scan System (2016). https://www.tekscan.com/products-solutions/systems/i-scan-system.
61. Pressure Profile Systems, RoboTouchTM Tactile Sensors with Digital Output for the Barrett Hand. https://static1.squarespace.com/static/53836bf1e4b011aa8ac0ffb9/t/53c85b3ae4b052122aacf107/1405639482642/PPS+_RoboTouch_SpecSheet.pdf.
62. A. J. Schmid, N. Gorges, D. Goger, H. Worn, "Opening a door with a humanoid robot using multi-sensory tactile feedback" in *2008 IEEE International Conference on Robotics and Automation* (IEEE, Pasadena, CA, USA, 2008; http://ieeexplore.ieee.org/document/4543222/), pp. 285–291.
63. Shadow Robot, Shadow Fingertip. http://www.shadowrobot.com/tactile/overview.shtml.
64. V. J. Lumelsky, E. Cheung, Real-time collision avoidance in teleoperated whole-sensitive robot arm manipulators. *IEEE Trans. Syst., Man, Cybern.* **23**, 194–203 (1993).

65. D. Um, B. Stankovic, K. Giles, T. Hammond, V. Lumelsky, "A modularized sensitive skin for motion planning in uncertain environments" in *Proceedings. 1998 IEEE International Conference on Robotics and Automation (Cat. No.98CH36146)* (IEEE, Leuven, Belgium, 1998; http://ieeexplore.ieee.org/document/676240/)vol. 1, pp. 7–12.
66. Y. Ohmura, Y. Kuniyoshi, "Humanoid robot which can lift a 30kg box by whole body contact and tactile feedback" in *2007 IEEE/RSJ International Conference on Intelligent Robots and Systems* (IEEE, San Diego, CA, USA, 2007; http://ieeexplore.ieee.org/document/4399592/), pp. 1136–1141.

# Supplementary Material for:

# Adaptive Compressive Tactile Subsampling: Enabling High Spatiotemporal Resolution in Scalable Robotic Skin

**Authors:** Ariel Slepyan[1]*†, Dian Li (李典)[2]†, Hongjun Cai[1], Ryan McGovern[2], Aidan Aug[2], Sriramana Sankar[2], Trac Tran[1], Nitish Thakor[1,2,3]*

**Affiliations:**

[1]Department of Electrical and Computer Engineering, Johns Hopkins University, 3400 North Charles Street, Baltimore, MD 21218, USA.

[2]Department of Biomedical Engineering, Johns Hopkins School of Medicine, 720 Rutland Avenue, Baltimore, MD 21205, USA.

[3]Department of Neurology, Johns Hopkins University, 600 North Wolfe, Baltimore, MD 21205, USA.

*Corresponding author. Ariel Slepyan: aslepya1@jhu.edu. Nitish Thakor: nitish@jhu.edu

†These authors contributed equally to this work

**Scalability of Tactile Sensor Arrays without integrated sensor electronics**

| | **Number of Sensors** | **Sampling Rate (Hz)** | **Total Sensor Rate (Sensors/Sec)** |
|---|---|---|---|
| Scalable tactile glove (***15***) | 548 | 7 | **3,836** |
| Conformal tactile textiles (***16***) | 216 | 14 | **3,024** |
| Gifu Hand III | 859 | 100 | **85,900** |
| TactArray system (Pressure Profile Systems) (***56***) | 1024 | 10 | **10,240** |
| Industrial I-scan system (TekScan) (***57***) | 1936 | 100 | **193,600** |
| RoboTouch$^{TM}$ (***58***) | 162 | 30 | **4,860** |
| Multi-sensory humanoid robot (***59***) | 504 | 40 | **20,160** |
| Shadow Robot (***60***) | 34 | 20 | **680** |
| Whole-sensitive robot arm (***61***) | 475 | 20 | **9,500** |
| Modularized sensitive skin (***62***) | 960 | 100 | **96,000** |
| Whole body contact humanoid robot (***63***) | 1864 | 20 | **37,280** |
| **Adaptive Compressive Tactile Subsampling (This work)** | 1024 | 1000 | **1,024,000** |

**Figure S1.** Table of large-area tactile sensor arrays without integrated sensor electronics with number of sensors, sampling rate, and the total sensor rate.

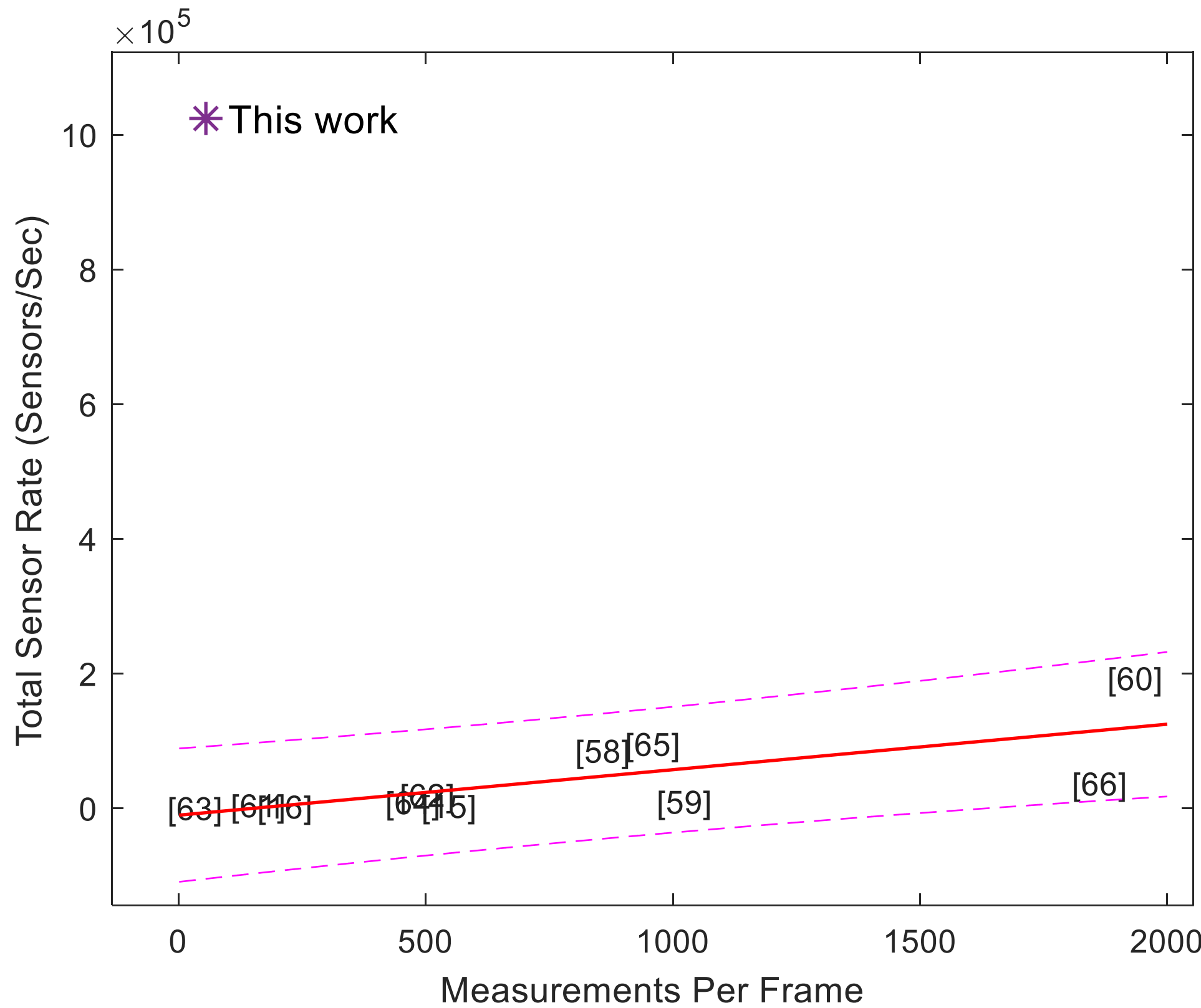

**Figure S2.** Total sensor rate versus number of measurements per frame for the sensor arrays in S1 and the compressive tactile subsampling sensor.

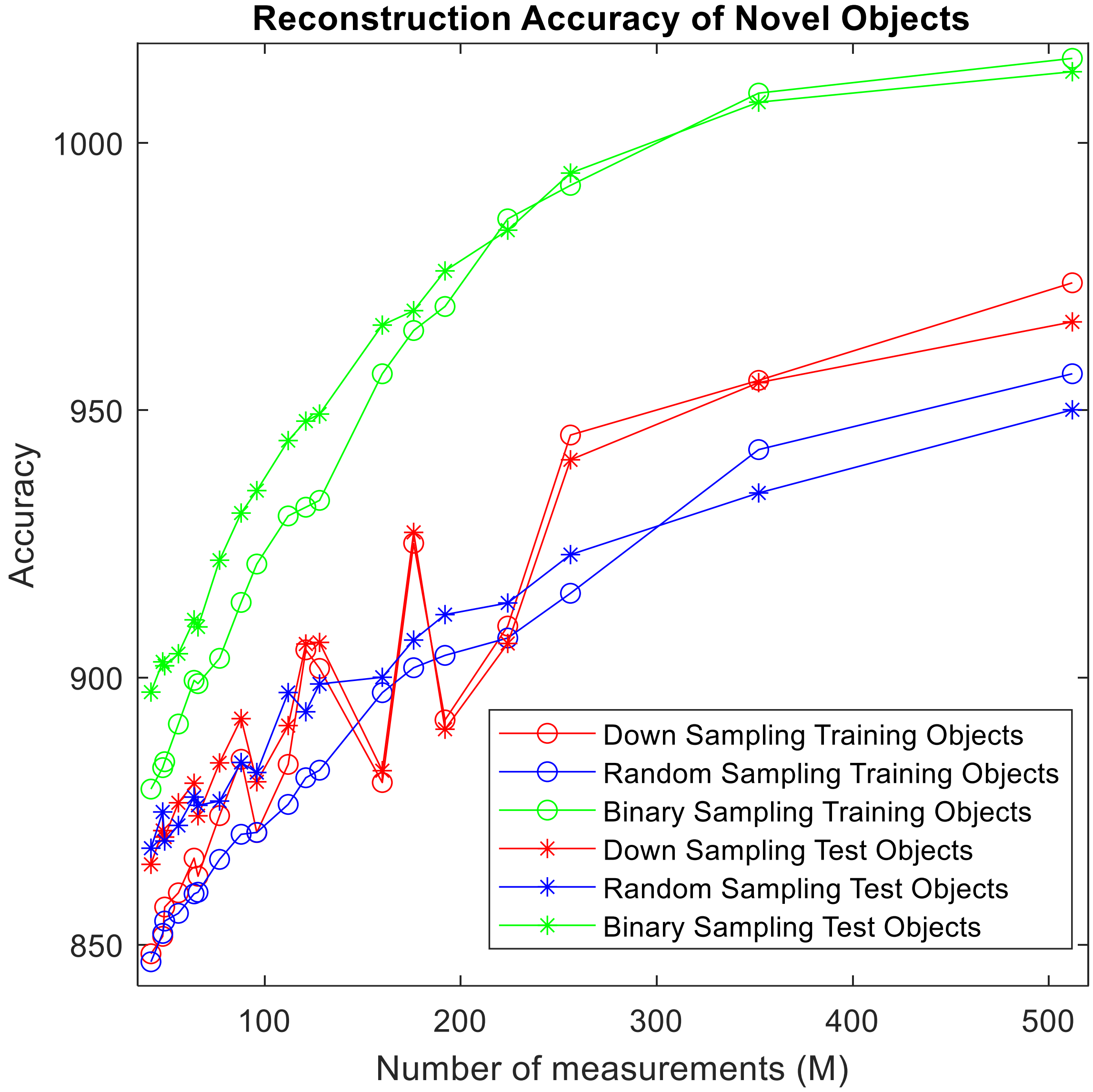

**Fig. S3-1. Generalizability of Compressive Subsampling to New Objects at different Measurement Levels.** Test objects were never seen in the dictionary learning process and achieve comparable accuracy across measurement levels and measurement schemes.

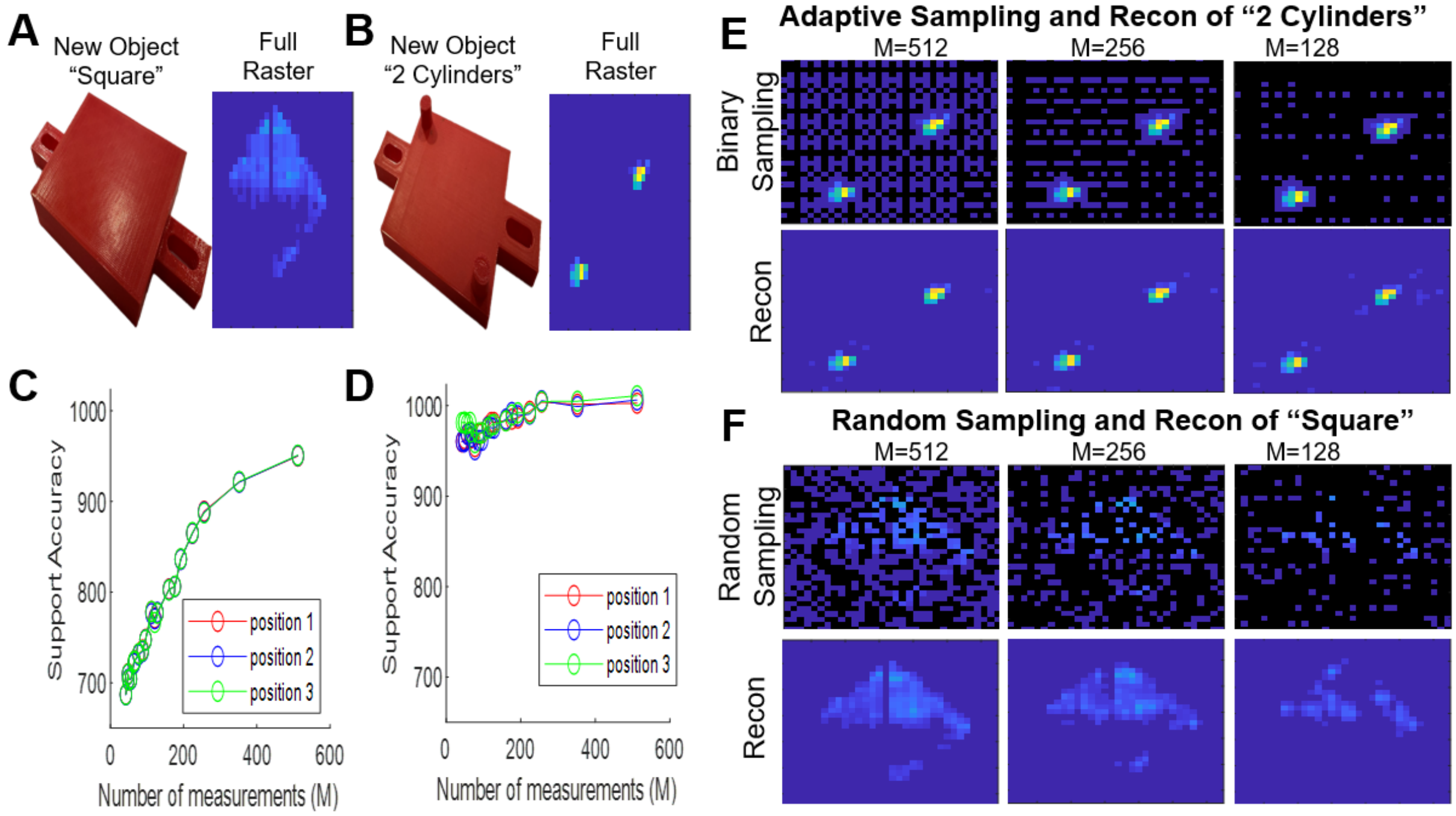

**Fig. S3-2. Generalizability of Compressive Subsampling to New Objects. (A)** Image and full raster scan measurement of a large new object not used in dictionary learning "Square". **(B)** Image and full raster scan measurement of a small new object not used in dictionary learning "2 Cylinders". **(C)** Support Accuracy vs Measurement Level of "Square" object pressed on the sensor at three different locations. **(D)** Support Accuracy vs Measurement Level of "2 Cylinders" object pressed on the sensor at three different locations. **(E)** Adaptive sampling and Reconstruction for "2 Cylinders" at location 1. **(F)** Random Sampling and Reconstruction for "Square" at location 1.

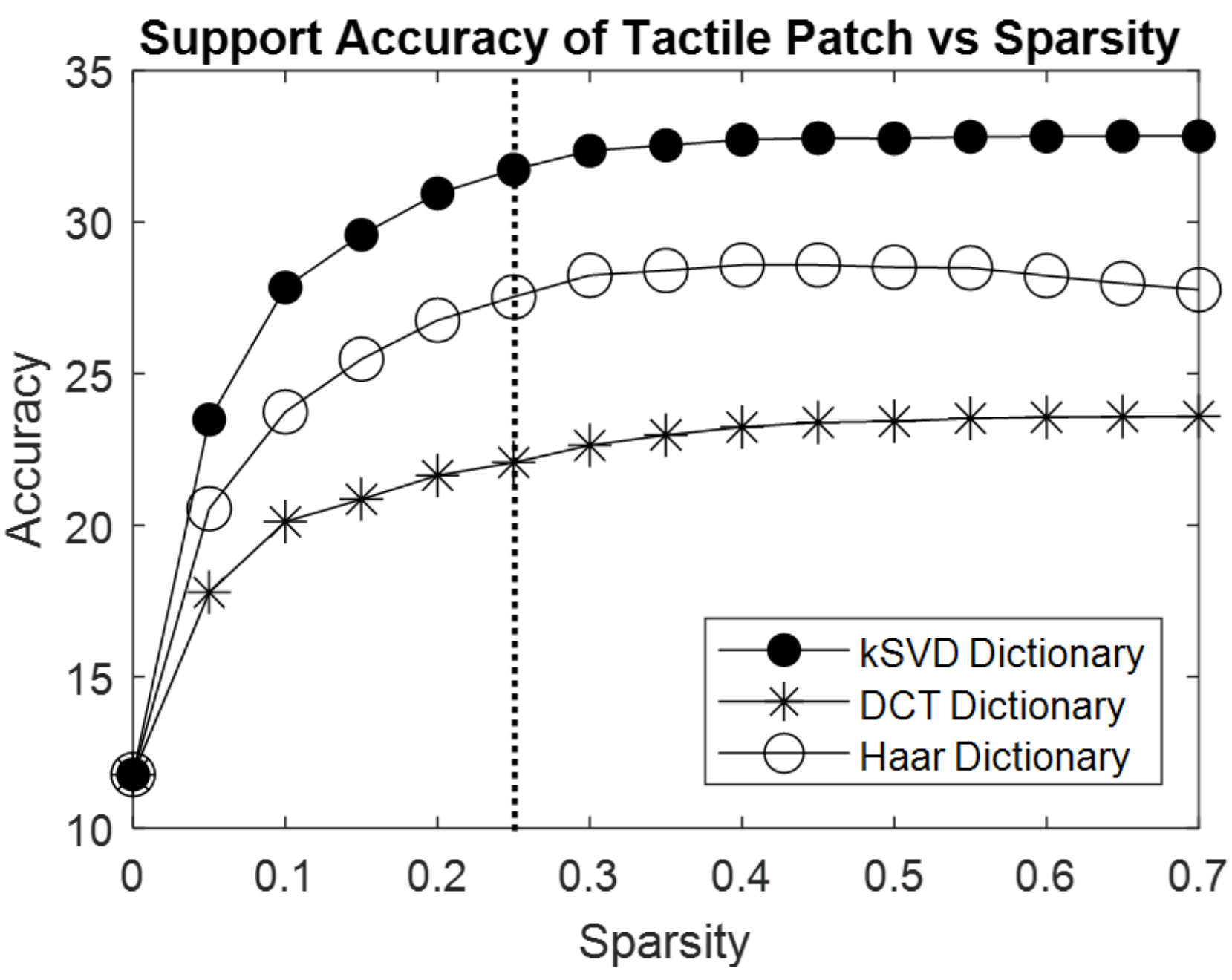

**Fig. S4. Effect of Sparsity Level on Reconstruction Accuracy.** Increasing sparsity level during tactile reconstruction has diminishing benefits on accuracy. 25% (dotted line) is used as the sparsity level as is typical in compressed sensing works.

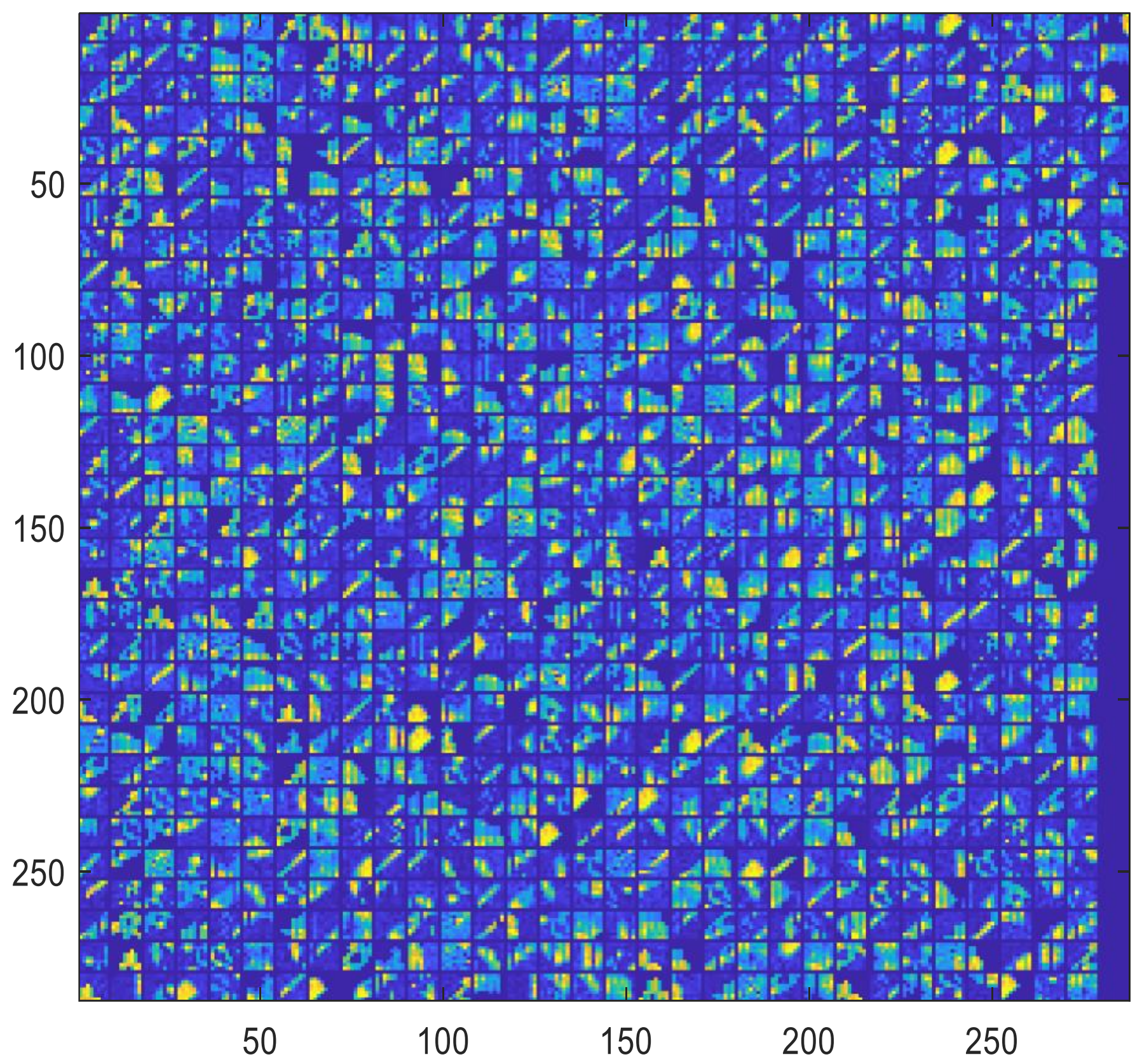

**Figure S5. Learned tactile dictionary used for recovery of subsampled tactile signals.** The dictionary is of size 64x1000 and is learned from the patch of size of 8x8 randomly selected from the full raster image of objects. Here, each small block represents each atom of the dictionary.

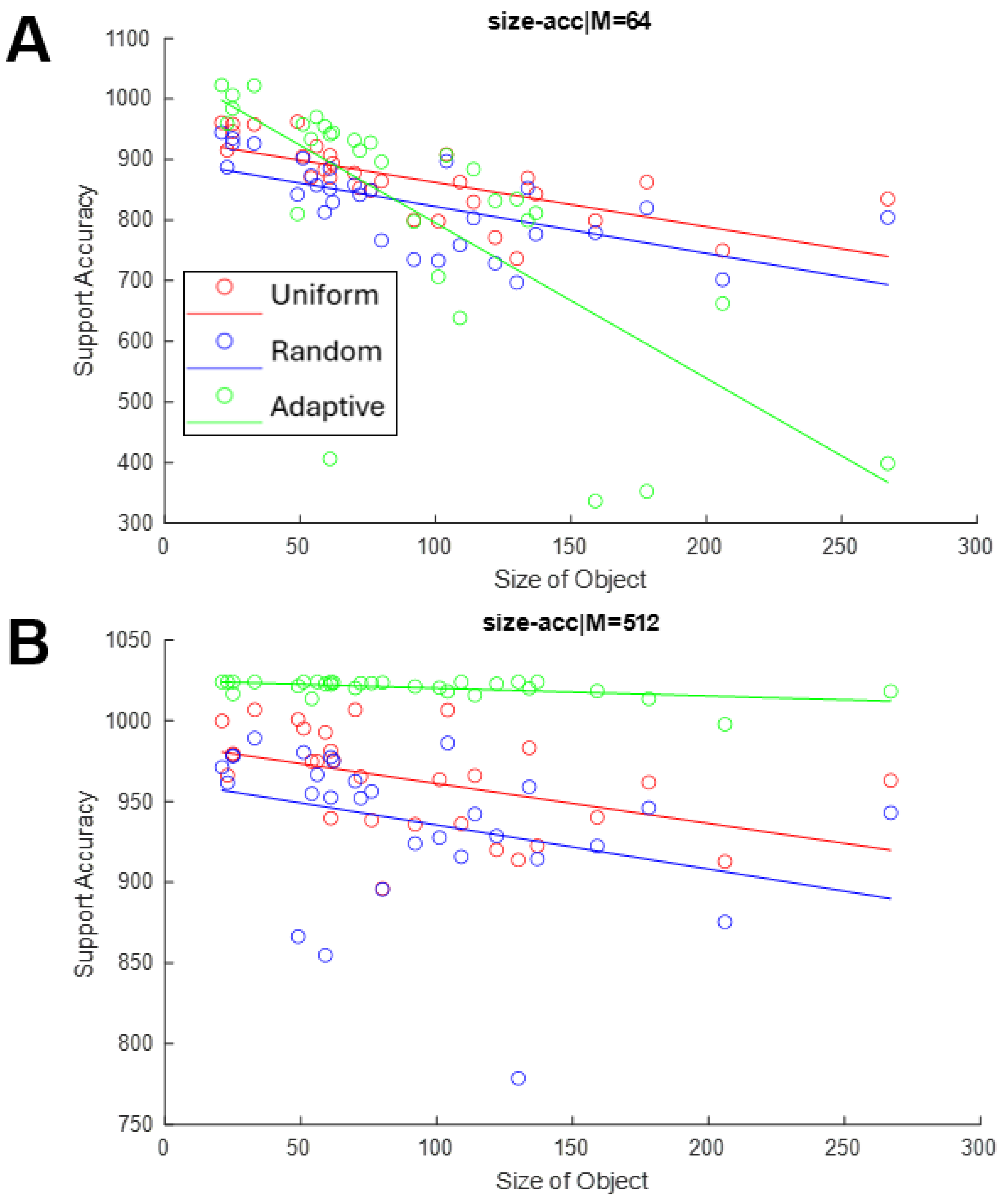

**Figure S6. Object Size vs Support Accuracy for M=64 and M=512. (A)** When the number of measurements is small (M=64), the size of the object impacts which subsampling method is most effective. **(B)** When the number of measurements is large (M=512), the adaptive, uniform, and random methods are best respectively.

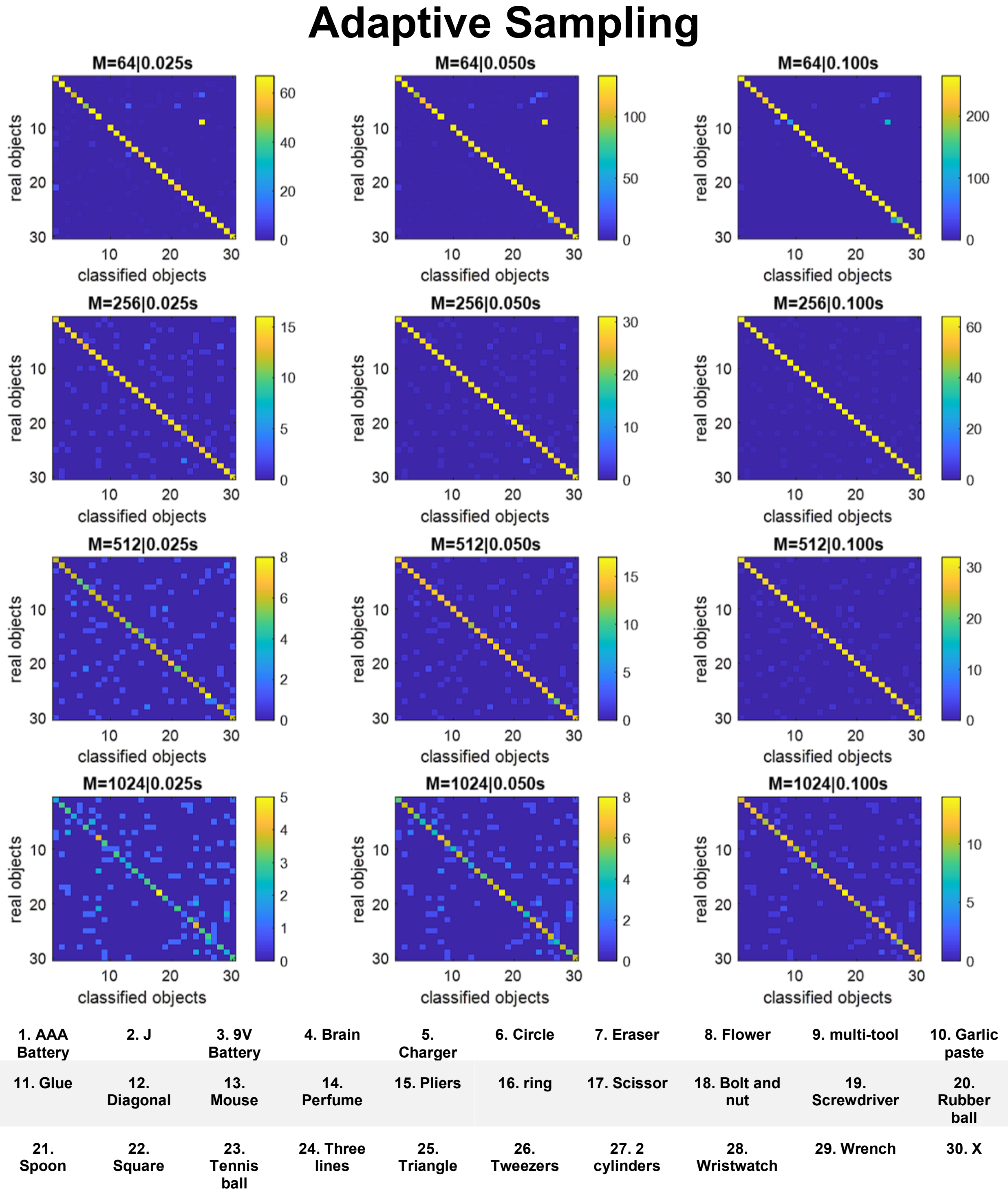

**Figure S7-1.** Confusion matrices for classification of objects using Adaptive sampling at different times since first contact.

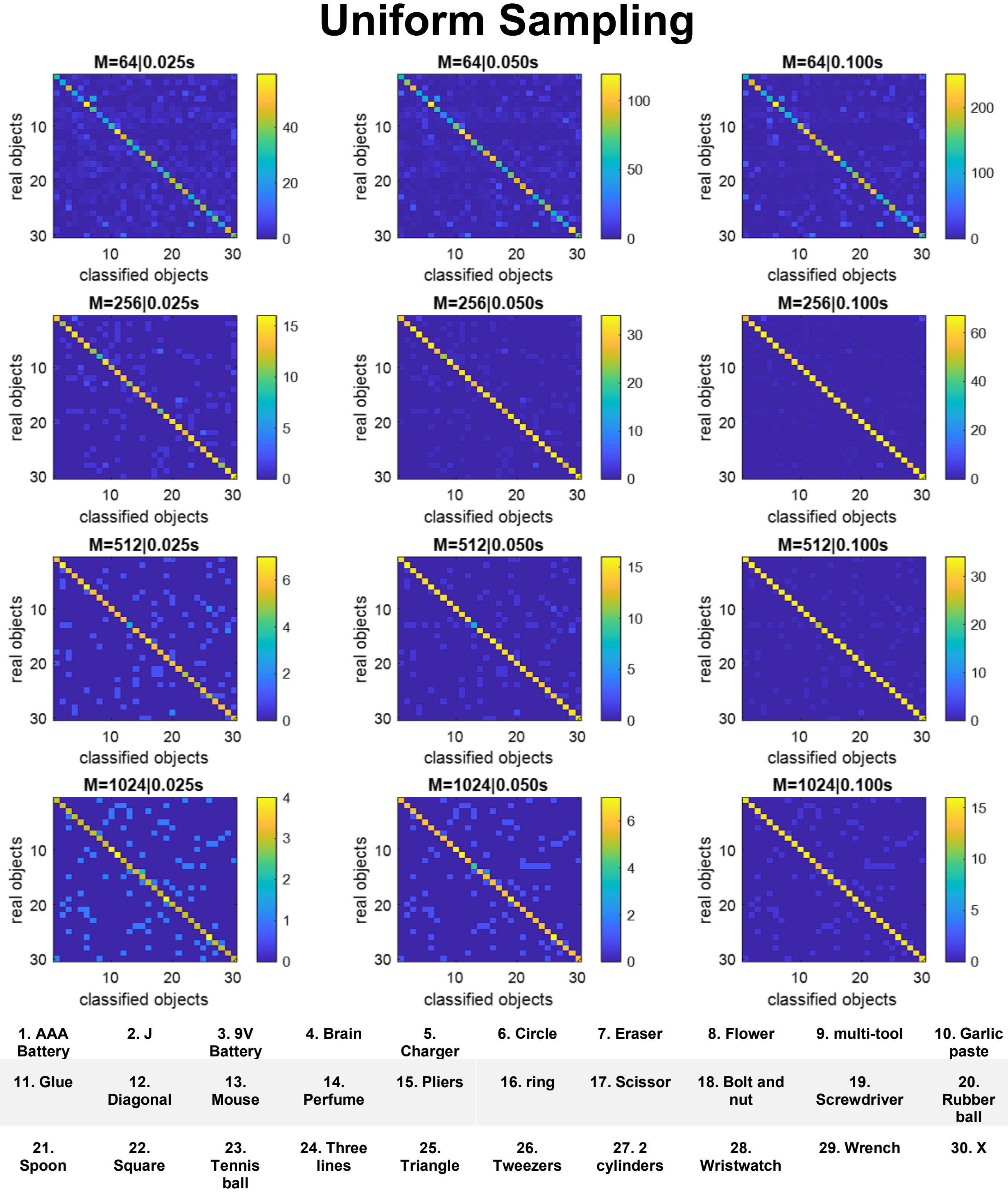

**Figure S7-2.** Confusion matrices for classification of objects using Uniform Sampling at different times since first contact.

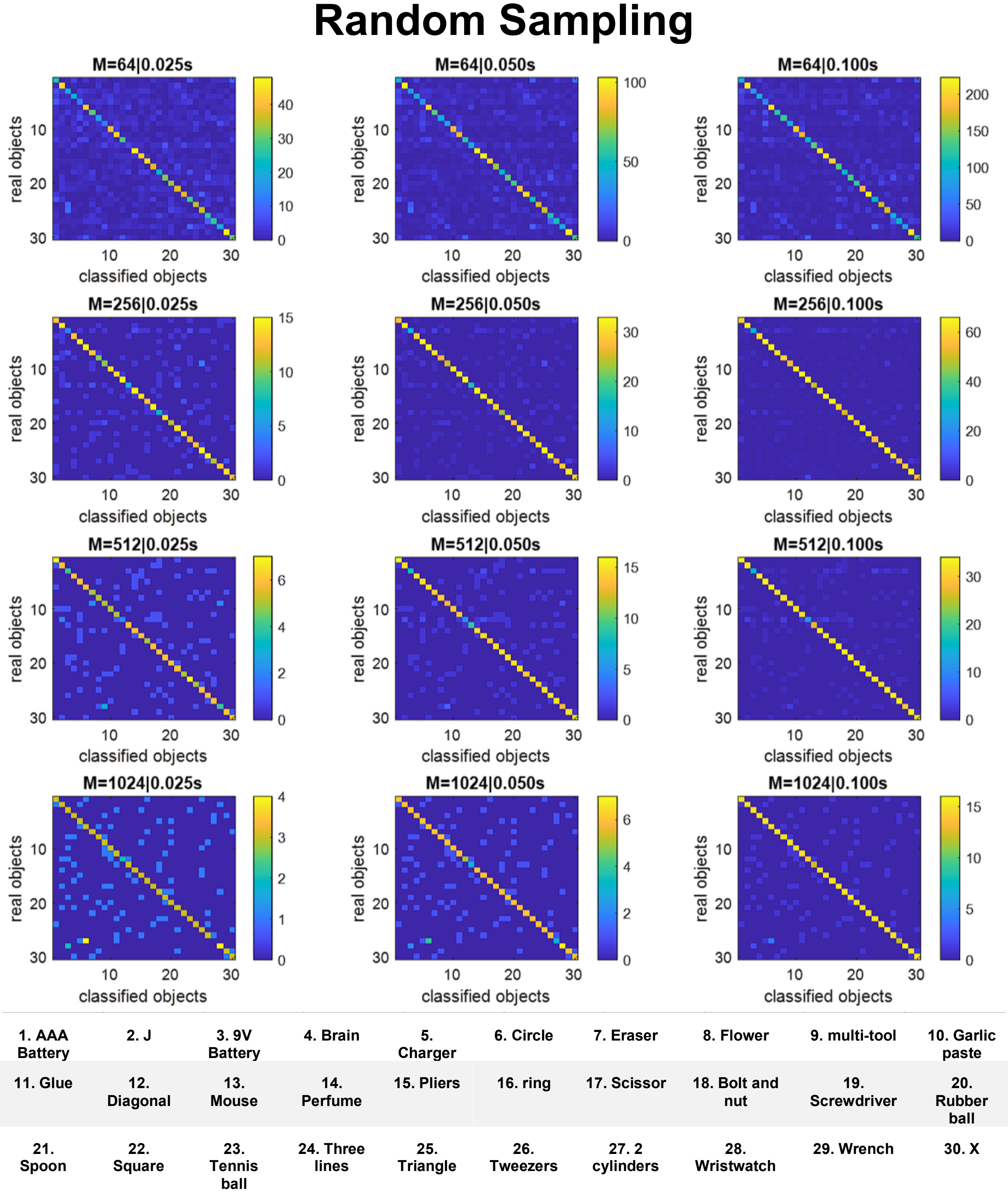

**Figure S7-3.** Confusion matrices for classification of objects using Random Sampling at different times since first contact.

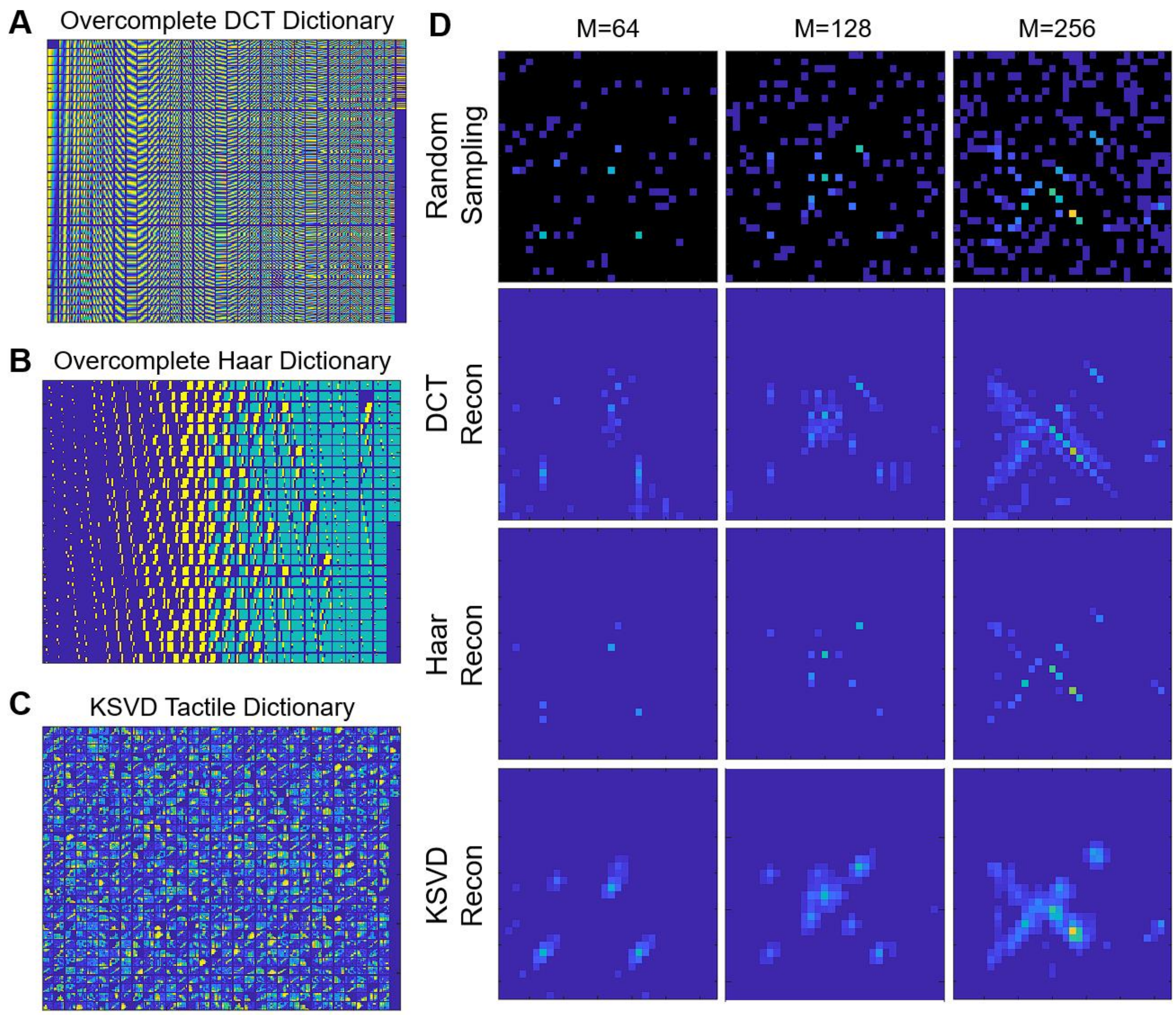

**Figure S8-1.** Effect of other dictionaries (Overcomplete Haar and Overcomplete DCT) on sensor reconstruction at different measurement levels.

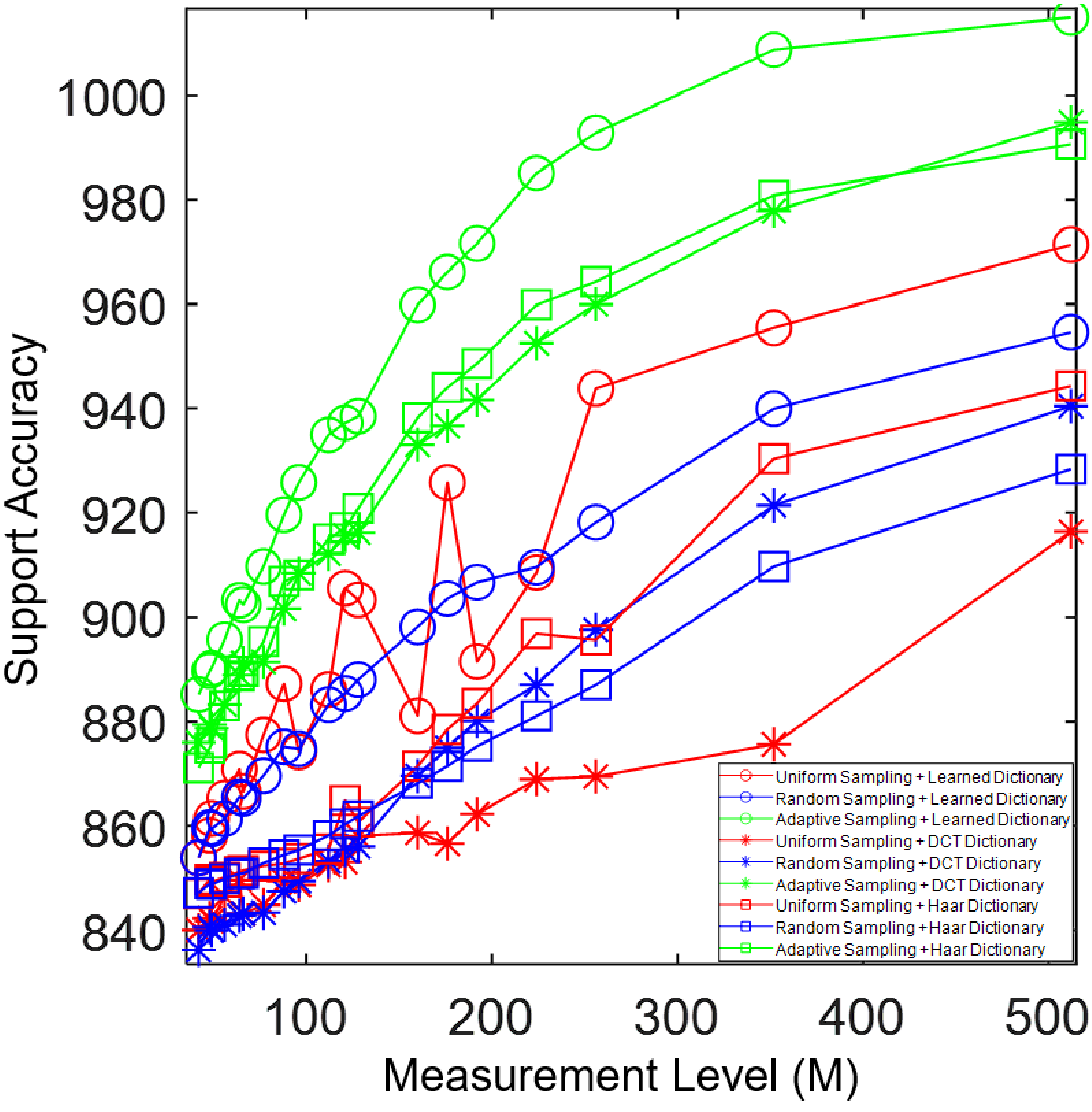

**Figure S8-2.** Effect of other dictionaries (Overcomplete Haar and Overcomplete DCT) on support accuracy at different measurement levels and sampling schemes.

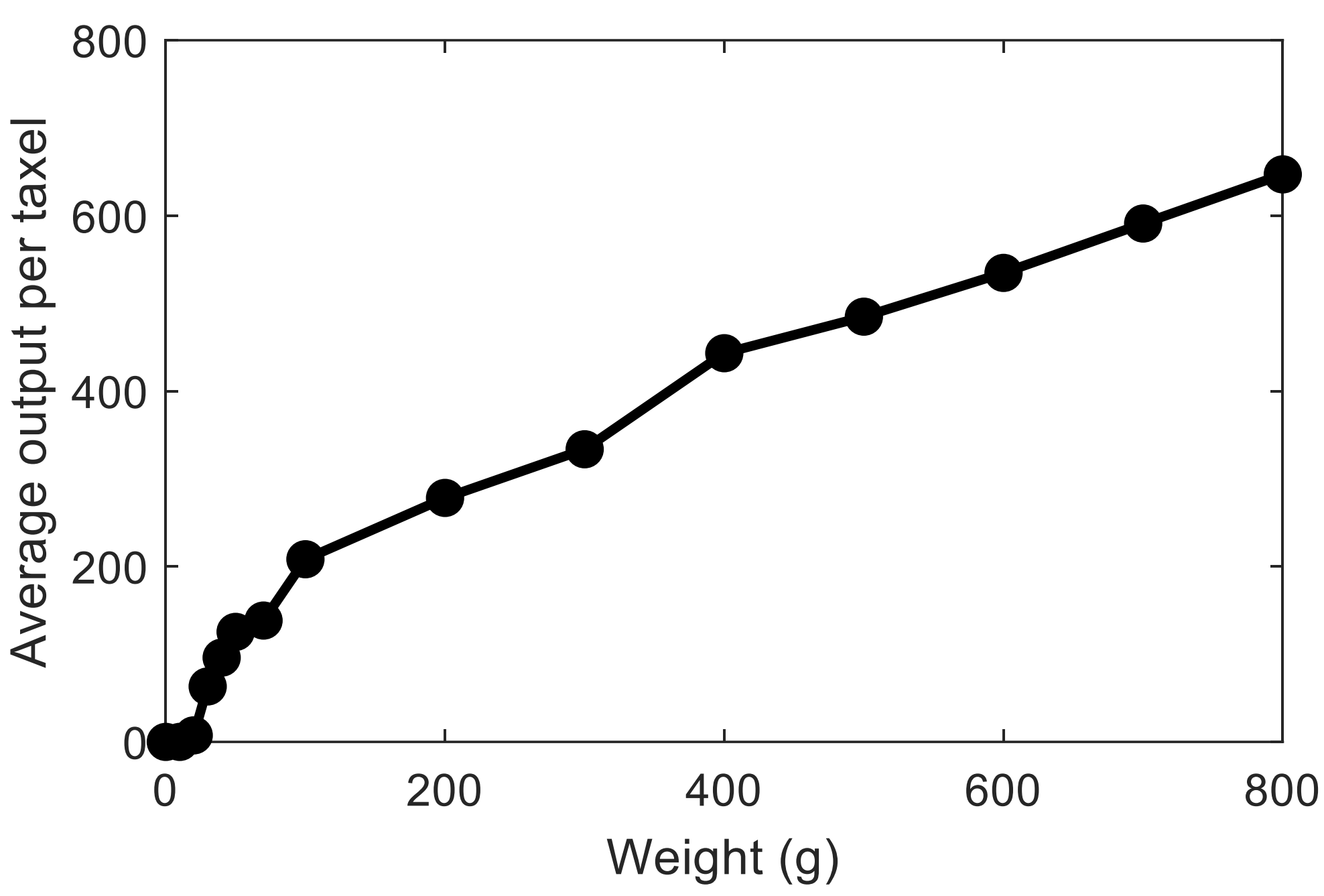

**Figure S9. Force response of each taxel on the sensor.**

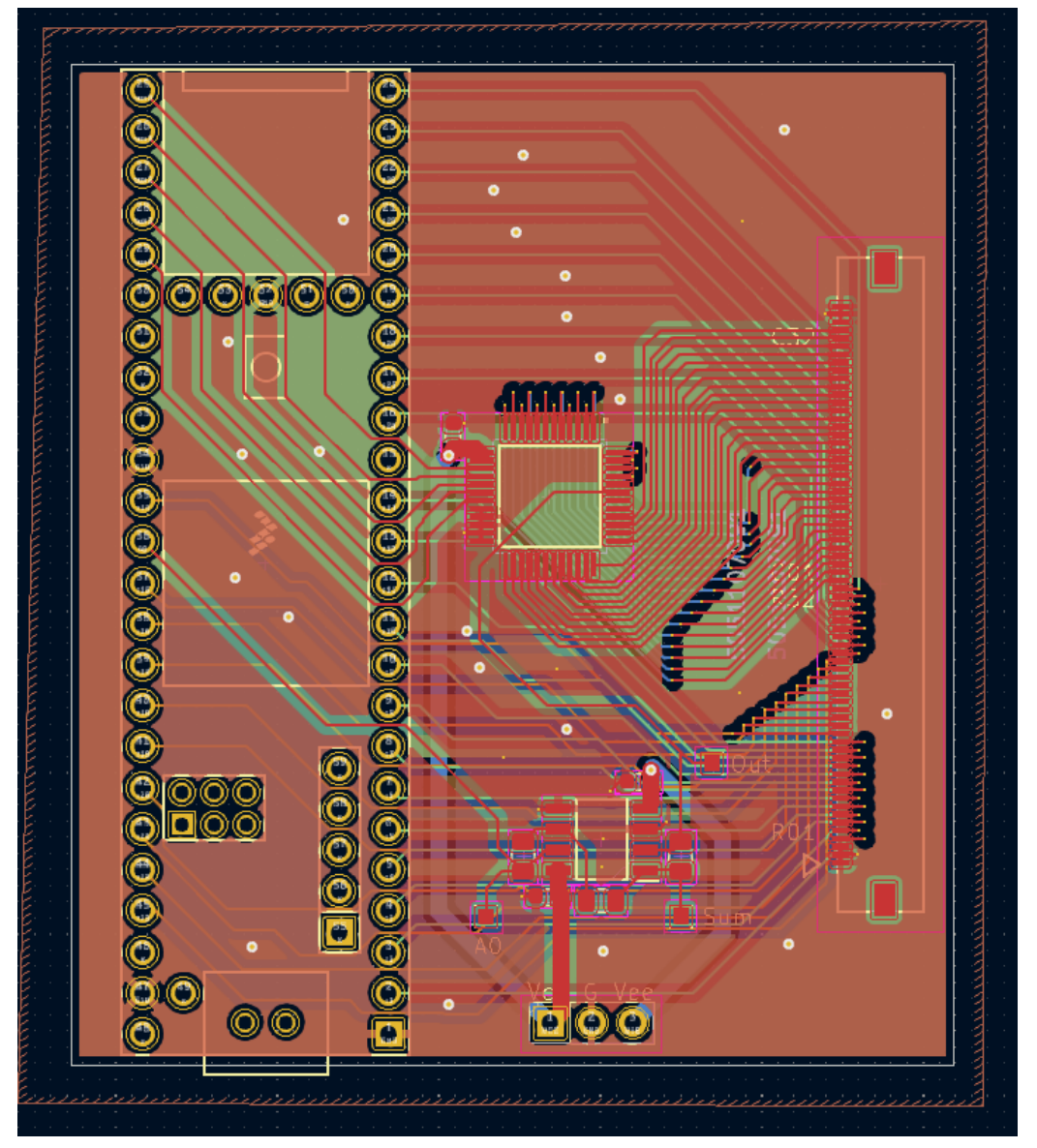

PCB Layout of Readout Board

PCB Layout of Sensor Array

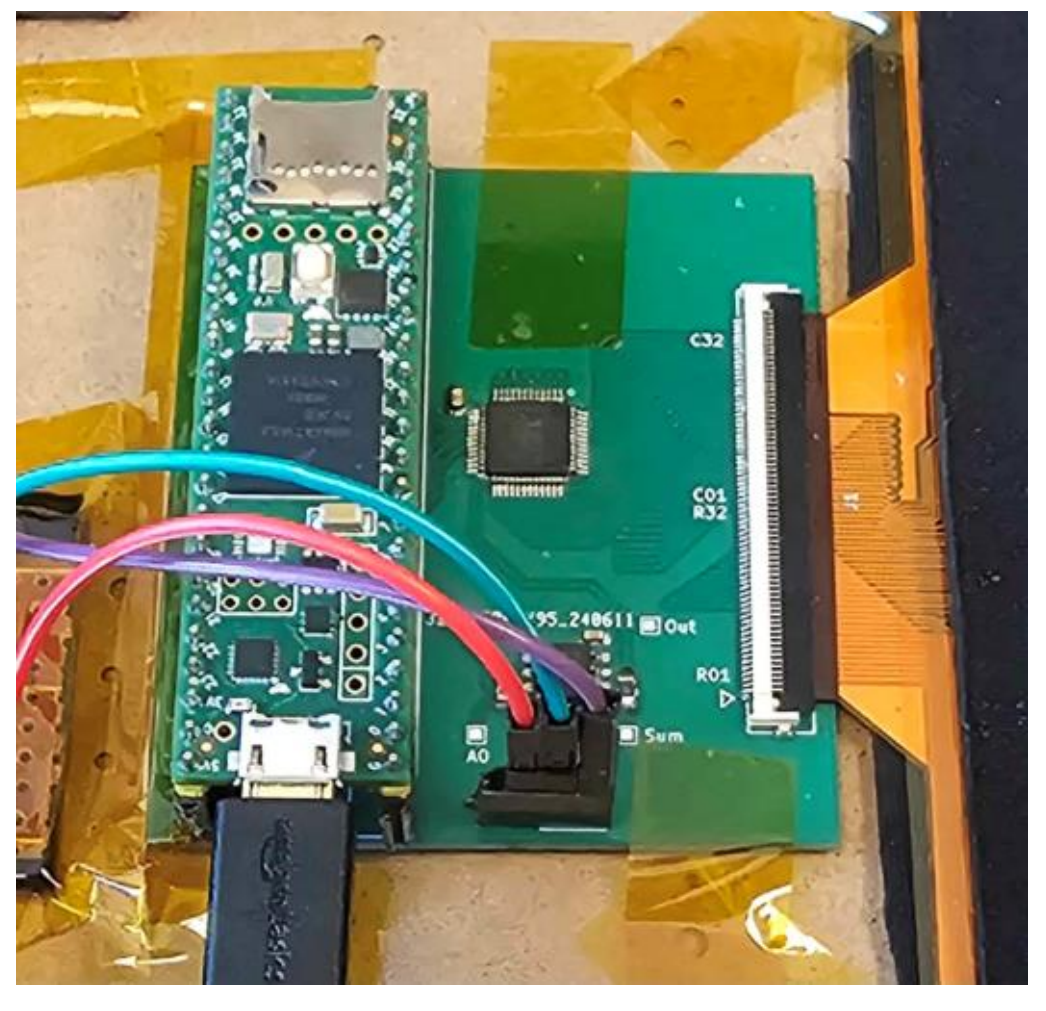

Image of Readout Board

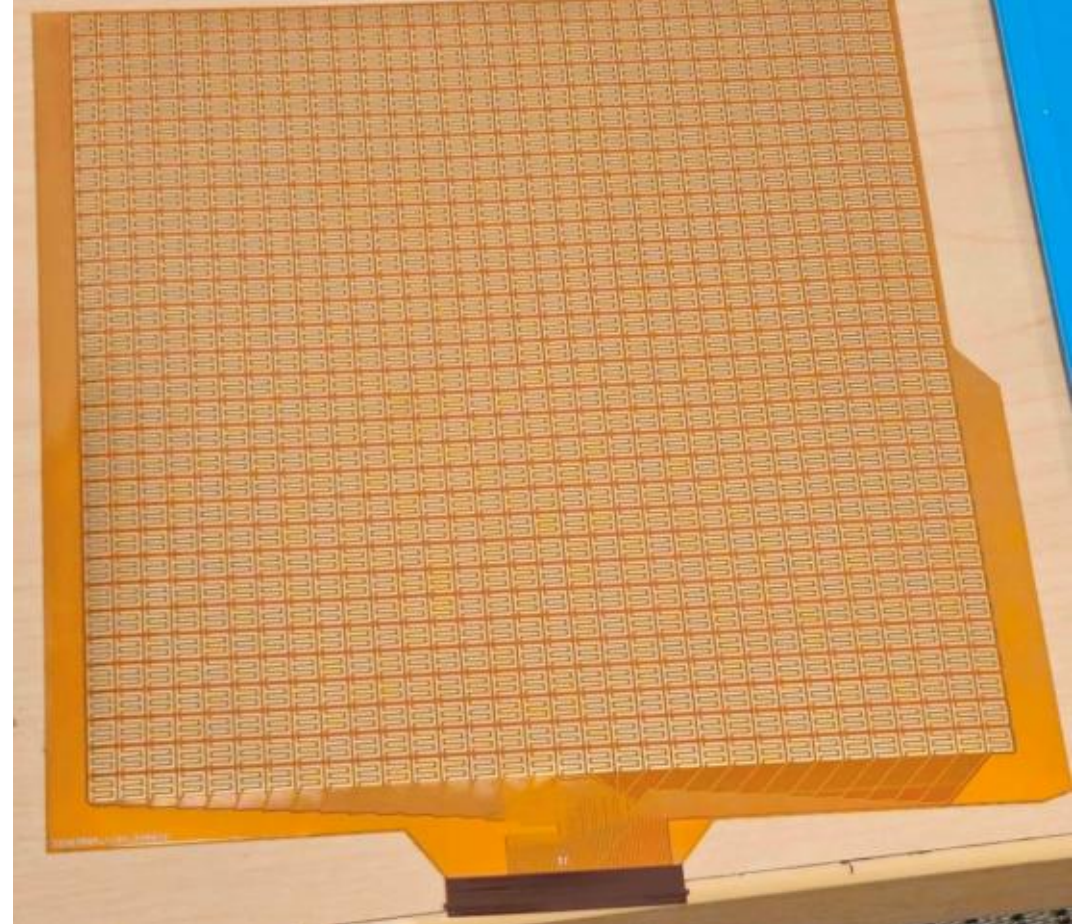

Image of Sensor Array

**Figure S10-1. Circuit board layout and images**

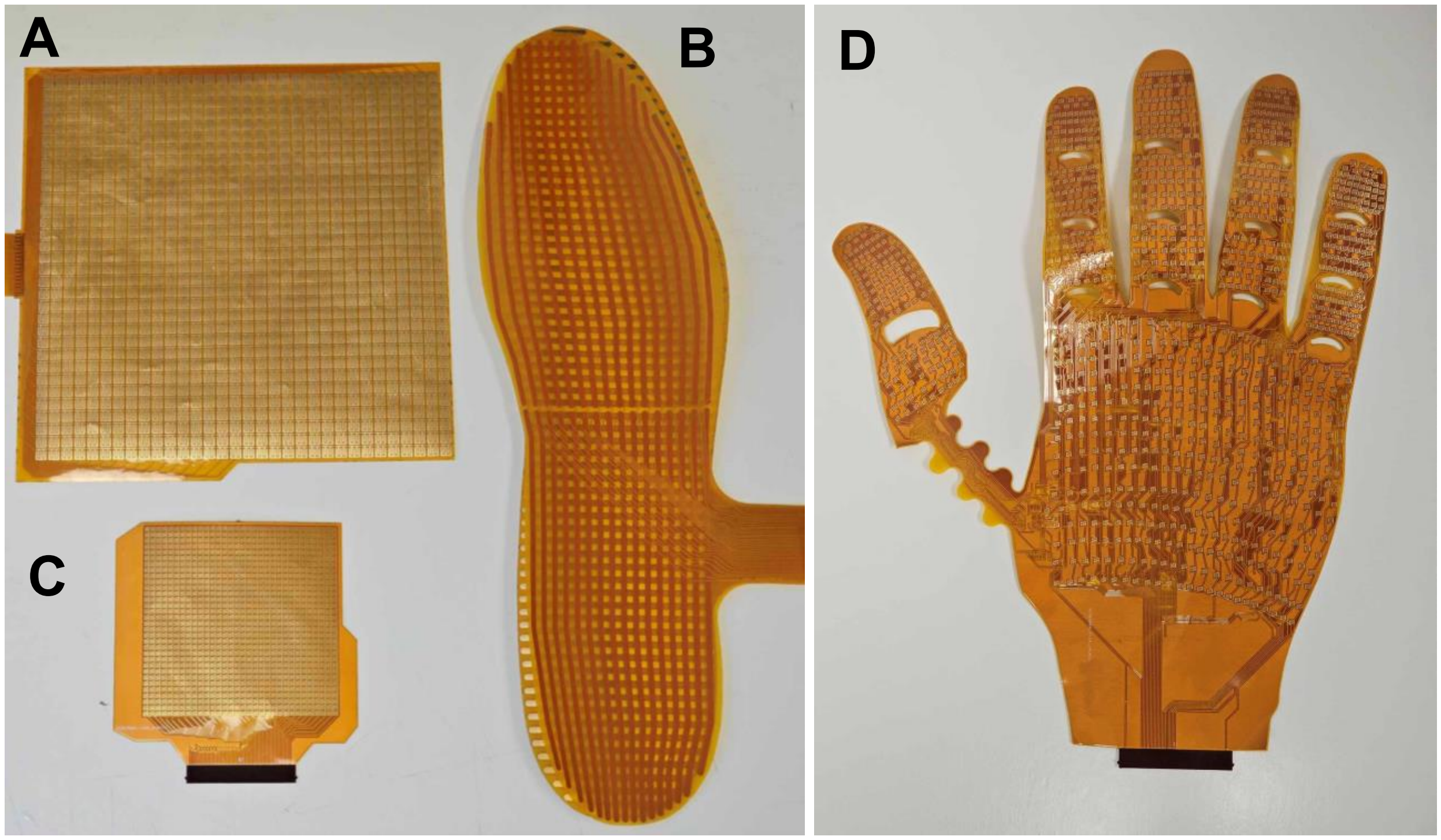

**Figure S10-2. Additional Sensor Arrays. (A)** Original sensor array 140mm x 140mm. **(B)** High-density sensor array 64mm x 64mm. **(C)** Insole sensor array 280mm x 85mm. **(D)** High-Density tactile sensing glove 230mm x 180mm.

| Reference | Value |
| --- | --- |
| C1, C3, C4 | 0.1µF, 0805 |
| IC1 | ADG732BSUZ |
| IC2 | TLV9362 |
| J2 | Conn_01x03_Pin, 2.54mm |
| J3 | 5051107091, 70-circuit FPC connector |
| R1 | 500 Ω, 0805 |
| R2, R3 | 1000 Ω, 0805 |
| U1 | Teensy 4.1 |
| Piezoresistive Layer | Velostat |
| Protective Foam | Neoprene |

**Figure S11. Readout board Component List**

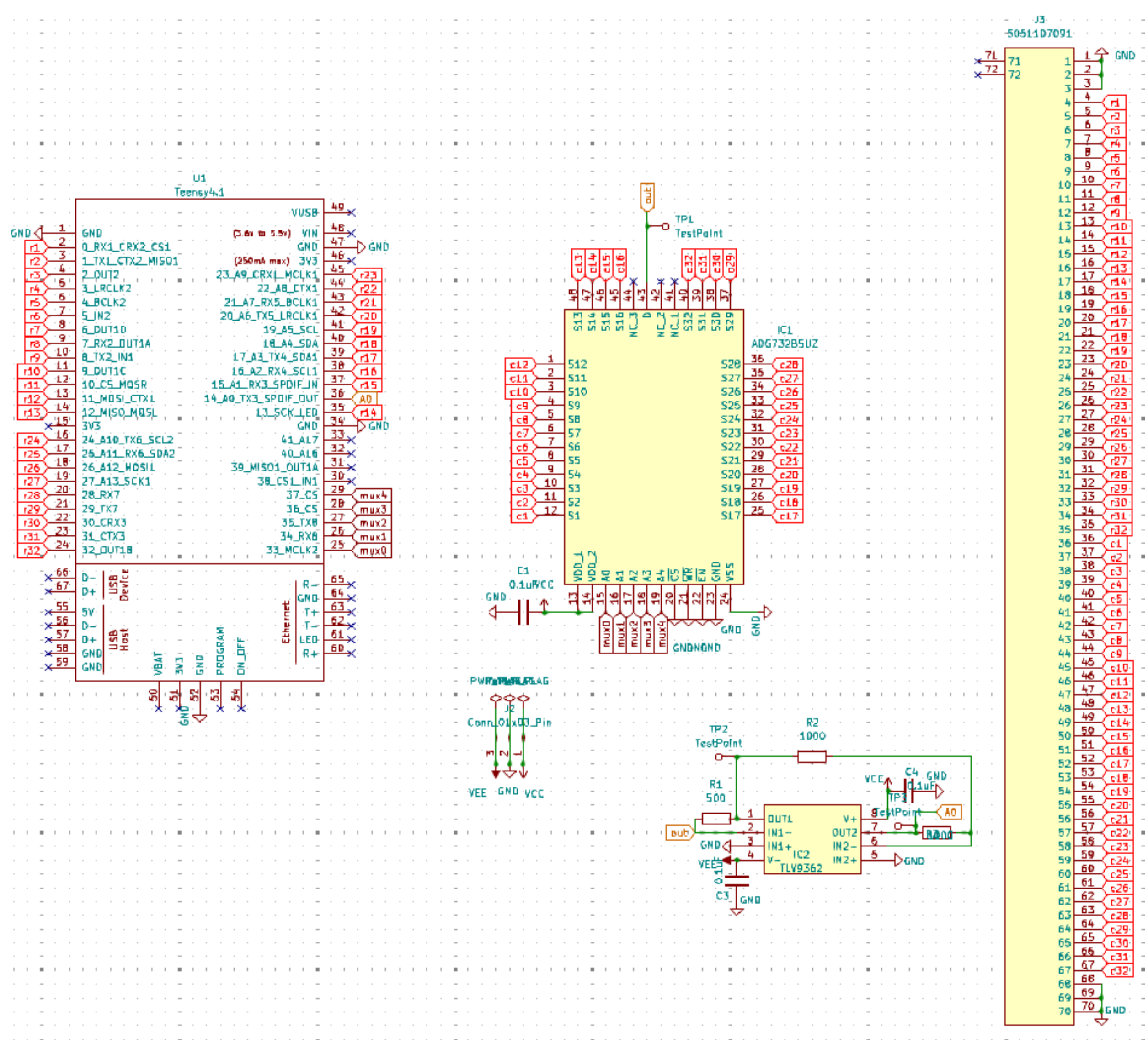

**Figure S12. Readout board schematic**

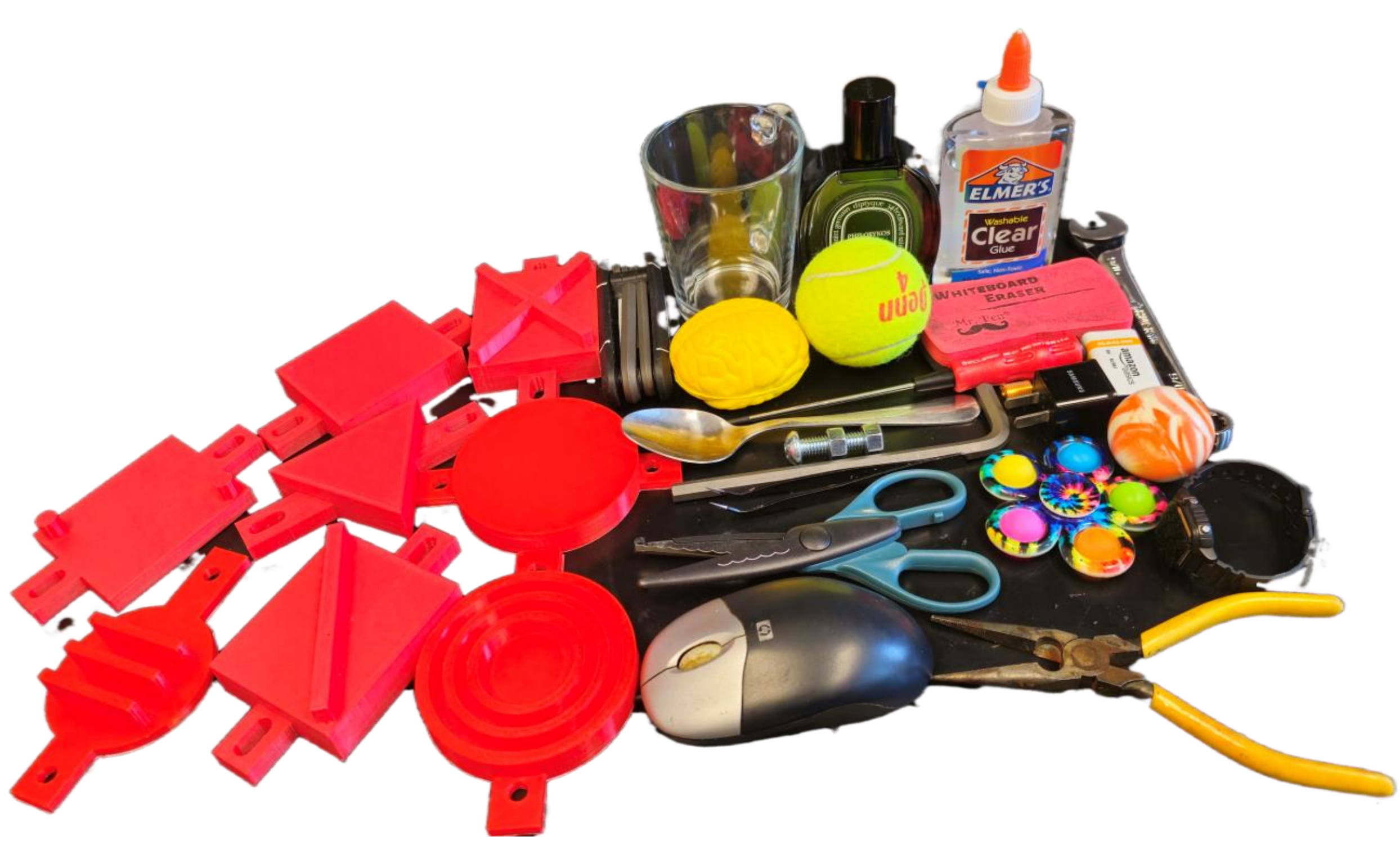

3D printed objects: (8 objects)

“2 cylinders”, “3 lines”, “diagonal line”, “3 rings”, “triangle”, “circle”, “square”, “X”.

Daily objects: (22 objects)

“computer mouse”, “art scissors”, “pliers”, “wristwatch”, “fidget spinner”, “rubber ball”, “hex bar”, “AAA battery”, “wrench”, “bolt and nut”, “spoon”, “soft brain”, “tennis ball”, “glass mug”, “multitool”, “cologne”, “glue bottle”, “9V battery”, “eraser”, “power adapter”, “screwdriver”, “tweezers”

**Figure S13. Picture of the 30 objects**

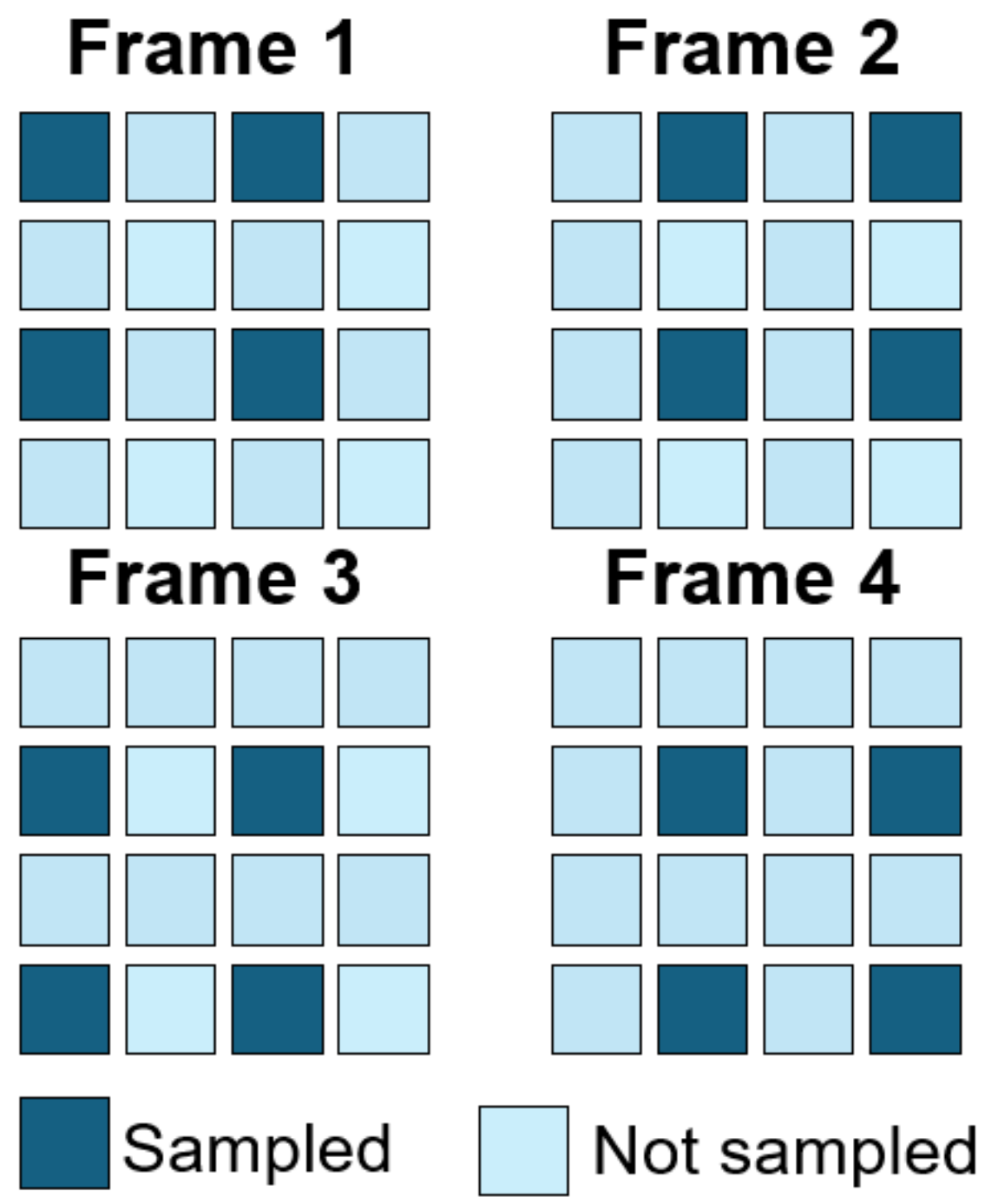

**Figure S14. Frame-wise shifting pattern of uniform subsampling method for a 4x4 sensor with M = 4.** After 4 frames, all pixels have been sampled.

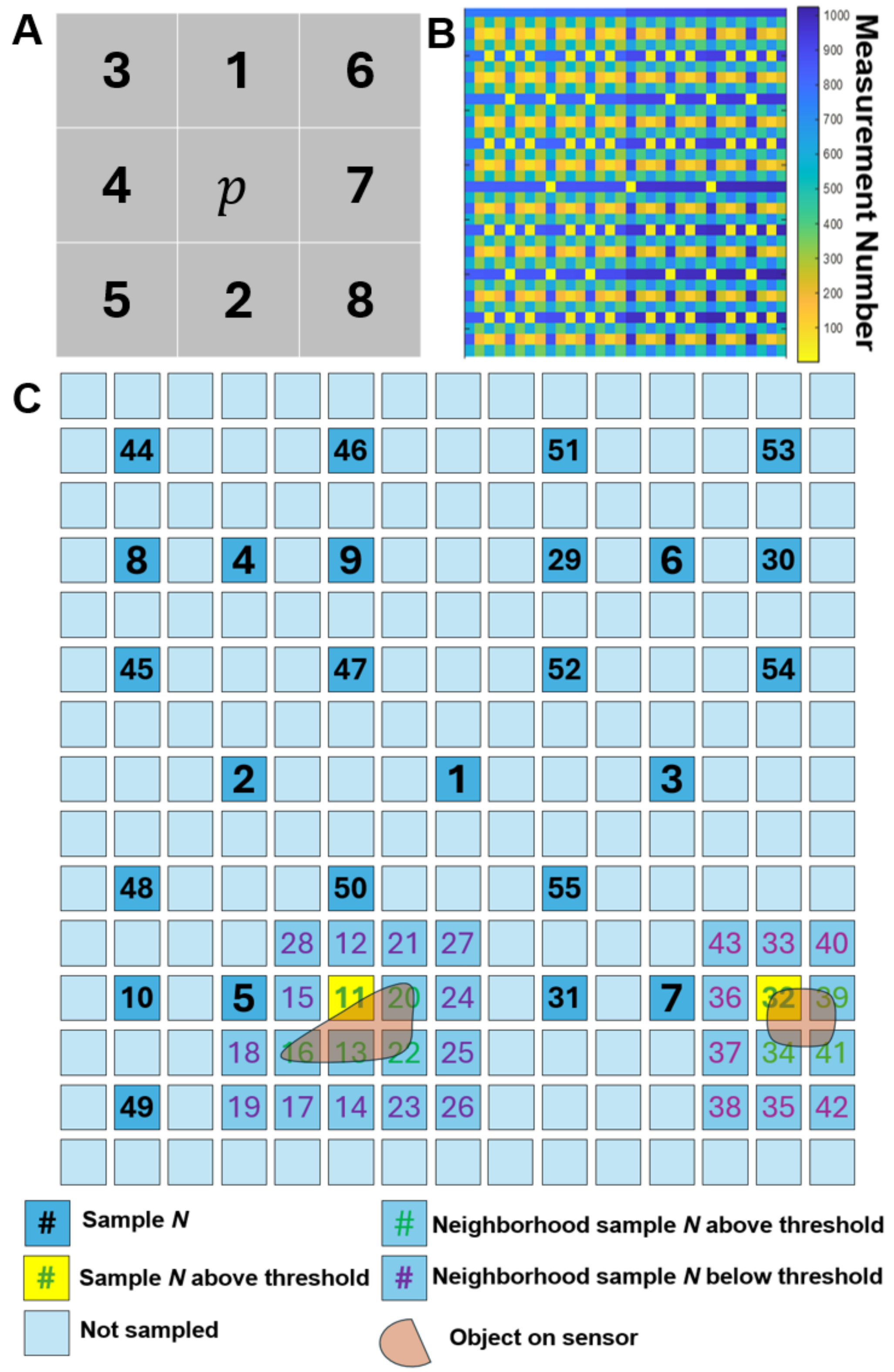

**Figure S15. Pattern of adaptive sampling with neighboring search.** (**A**) Order of neighboring search program around a point $p$ (NeighborList($p$)). (**B**) Full sampling pattern of the 32x32 square sensor for adaptive subsampling when no touches are detected. (**C**) Sampling pattern for adaptive subsampling with M = 55 and two small, detected objects on an example 15x15 sensor.

## Description of robotic tactile control

The system was implemented in ROS2 Humble using Universal Robots' official package which provides the robotic model, driver for UR5e. Our custom architecture extended this foundation with three specialized nodes:

- System state node: Processed raw tactile data (1000 Hz) to compute the ball's planar position by identifying the taxel with maximum pressure, publishing filtered states at 30 Hz.
- Tactile trajectory publisher: Generated reference trajectories, logged sensor data, and managed experiment sequencing.
- Tactile control node: Received estimated position and trajectory, computed tilt commands using PID controller, and dispatched joint angles to the UR5e controller.

The system dynamics were modeled using a discrete-time linear state-space system. The ball's position is measured based on the position of tactile sensors detecting maximum pressure. The discrete dynamic model for the system is described by the following state space equations:

$$s_{k+1} = \begin{bmatrix} 1 & 0 & \Delta t & 0 \\ 0 & 1 & 0 & \Delta t \\ 0 & 0 & 1 & 0 \\ 0 & 0 & 0 & 1 \end{bmatrix} s_k + \begin{bmatrix} 0 & 0 \\ 0 & 0 \\ g\Delta t & 0 \\ 0 & g\Delta t \end{bmatrix} \theta_k$$

Here, $s_k = [x_k \quad y_k \quad v_{x,k} \quad v_{y,k}]^T$ represents the state vector containing the planar position $(x, y)$ and velocity $(v_x \; v_y)$ of the ball at discrete time index $k$. Input vector, $\theta_k = [sin\theta_{1,k} \quad sin\theta_{2,k}]^T$ contains the sine values of the small tilt angles about the plate's $x$- and $y$-axes, measured with respect to the home (perfectly level) orientation. With sample period $\Delta t$ and gravity $g \approx 9.81 \; m/s^2$, the in-plane acceleration is calculated using $a = gsin\theta$.

We evaluated the performance of the system on both static and dynamic tasks. In one set of trials, the robot maintained the ball in a balanced position at the center of the array, despite external perturbations. In another, the system guided the ball along trajectories tracing the letters "N", "B", and "I", demonstrating accurate and repeatable motion control. Cumulative pressure maps for these letter trajectories are shown in Fig. 10, validating the system's ability to perform closed-loop control with only tactile feedback.

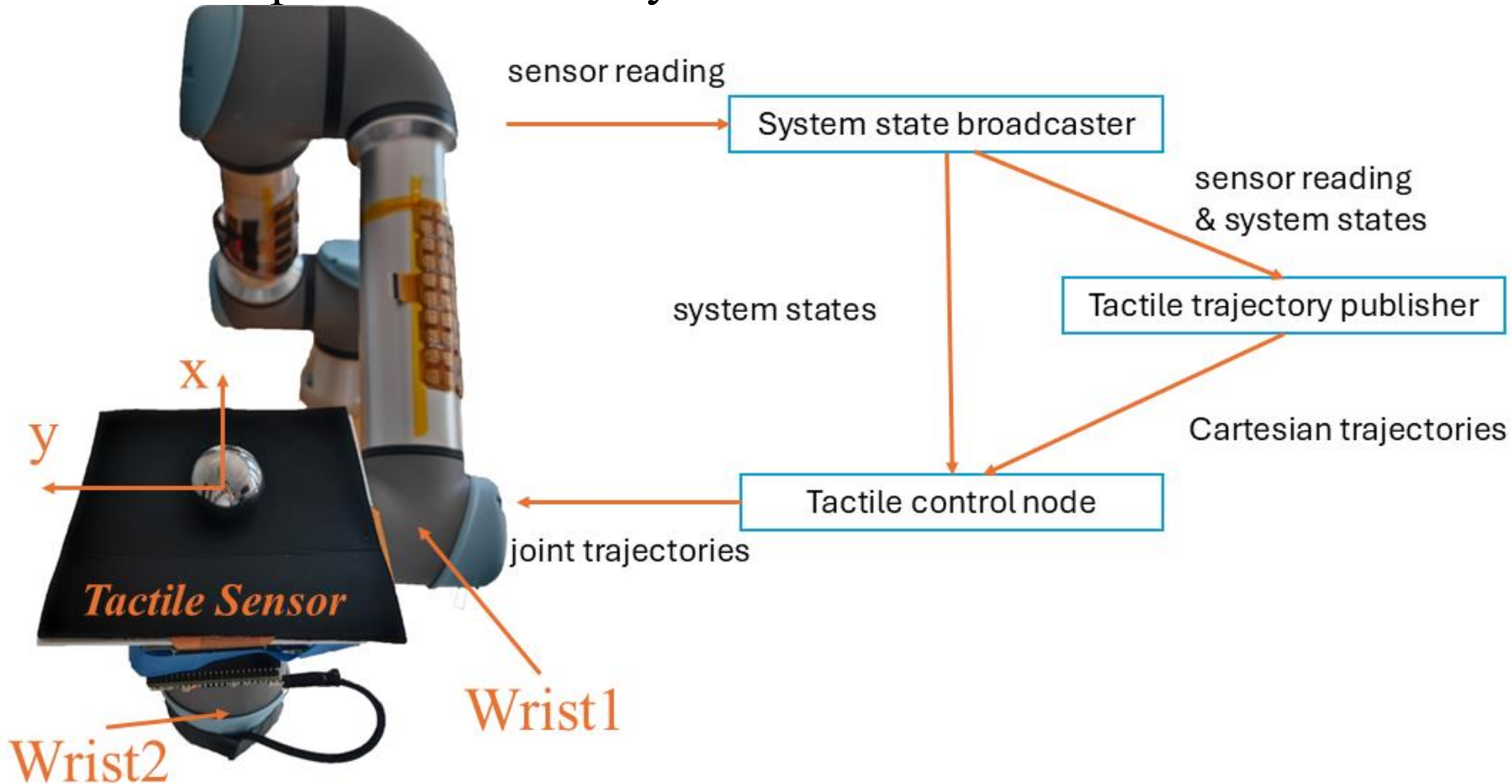

**Figure S16. Overview of the robotic control platform**

**Supplemental Algorithm 1. Adaptive sampling for a Single Frame**

1 **global** $\mathbf{PositionArray}[N]$, $\mathbf{ForceArray}[N]$, $i$

2 **def** AdaptiveSampling$(N, M, NSThr)$:

3 $\mathbf{BinaryOrderArray} \leftarrow \text{BuildBinaryOrderArray}(N)$;

4 $i \leftarrow 0$;

5 **for** $i^* \in \{1, \dots, N\}$

6 $p \leftarrow \mathbf{BinaryOrderArray}[i^*]$;

7 **if** ($p$ has been added in $\mathbf{PositionArray}$) **then** continue;

8 $force \leftarrow \text{ForceAt}(p)$;

9 $\mathbf{PositionArray}[i] \leftarrow p$; $\mathbf{ForceArray}[i] \leftarrow force$; $i \leftarrow i + 1$;

10 **if** $(force > NSThr)$ **then** NeighborSampling$(p, M, NSThr)$;

11 **if** $(i > M)$, **return** $\mathbf{PositionArray}$, $\mathbf{ForceArray}$;

12 **def** BuildBinaryOrderArray$(N)$:

13 $j \leftarrow 1$;

14 $\mathbf{CurrentCenterSet} \leftarrow \emptyset$; $\mathbf{LastCenterSet} \leftarrow \emptyset$;

15 $HorizontalDivision \leftarrow false$; $DivisionDistance \leftarrow \left\lceil \frac{\sqrt{N}}{2} \right\rceil$;

16 $\mathbf{BinaryOrderArray}[j] \leftarrow \left( \left\lceil \frac{\sqrt{N}}{2} \right\rceil, \left\lceil \frac{\sqrt{N}}{2} \right\rceil \right)$; $j \leftarrow j + 1$;

17 $\mathbf{LastCenterSet} \leftarrow \left( \left\lceil \frac{\sqrt{N}}{2} \right\rceil, \left\lceil \frac{\sqrt{N}}{2} \right\rceil \right)$;

18 **while** $(j <= N)$

19 **if** $(HorizontalDivision)$ **then**

20 **for** $(x, y) \in \mathbf{LastCenterSet}$

21 $\mathbf{CurrentCenterSet} \leftarrow \mathbf{CurrentCenterSet} \cup$

22 $\{(x - DivisionDistance, y), (x + DivisionDistance, y)\}$;

23 **if** ($(x - DivisionDistance, y)$has not been added in $\mathbf{BinaryOrderArray}$) **then**

24 $\mathbf{BinaryOrderArray}[j] \leftarrow (x - DivisionDistance, y)$; $j \leftarrow j + 1$;

25 **if** ($(x + DivisionDistance, y)$ has not been added in $\mathbf{BinaryOrderArray}$) **then**

26 $\mathbf{BinaryOrderArray}[j] \leftarrow (x + DivisionDistance, y)$; $j \leftarrow j + 1$;

27 **else**

28 $DivisionDistance \leftarrow \left\lceil \frac{DivisionDistance}{2} \right\rceil$;

29 **for** $(x, y) \in \mathbf{LastCenterSet}$

30 $\mathbf{CurrentCenterSet} \leftarrow \mathbf{CurrentCenterSet} \cup$

31 $\{(x, y - DivisionDistance), (x, y + DivisionDistance)\}$;

32 **if** (($x, y - DivisionDistance$) has not been added in **BinaryOrderArray**) **then**

33 **BinaryOrderArray**[$j$] ← ($x, y - DivisionDistance$); $j \leftarrow j + 1$;

34 **if** (($x, y + DivisionDistance$) has not been added in **BinaryOrderArray**) **then**

35 **BinaryOrderArray**[$j$] ← ($x, y + DivisionDistance$); $j \leftarrow j + 1$;

36 **LastCenterSet** ← **CurrentCenterSet**; **CurrentCenterSet** ← ∅;

37 $HorizontalDivision \leftarrow HorizontalDivision = false$;

38 **return BinaryOrderArray**;

39 **def** NeighborSampling($p', M, NSThr$):

40 **if** ($i > M$), **return**;

41 **for** $p'' \in$ NeighborList($p'$)

42 **if** ($p''$ has been added in **PositionArray**) **then** continue;

43 $force$ ← ForceAt($p''$);

44 **PositionArray**[$i$] ← $p''$; **ForceArray**[$i$] ← $force$; $i \leftarrow i + 1$;

45 **if** ($force > NSThr$) **then** NeighborSampling($p'', M, NSThr$)